%% file: main.tex
\documentclass[3p]{elsarticle}

\input{structure}

\begin{document}

\input{frontmatter}

\section{Introduction}
\subimport*{1_introduction/}{main}

\section{Materials and method}%
\label{sec:materials_and_method}
\subimport*{2_methods/}{main}

\section{Generation of pseudo-CT images}%
\label{sec:generation_pseudo}
\subimport*{3_unet/}{main}

\section{Instance segmentation using simulated images}%
\label{sec:instance_segmentation}
\subimport*{4_detectron/}{main}

\section{Textile model generation}%
\label{sec:mesh_generation}
\subimport*{5_meshing/}{main}

\section{Validation of textile model}%
\label{sec:validation}
\subimport*{6_validation/}{main}

\section{Conclusions and perspectives}%
\label{sec:conclusions}
\subimport*{7_conclusions/}{main}

\biboptions{sort}
\bibliography{main}

\appendix
\subimport*{8_appendix/}{main}

\end{document}

%% file: structure.tex
\usepackage{graphicx} 

\usepackage{import} 

\usepackage[utf8]{inputenc} 

\usepackage{textcomp} 

\usepackage[T1]{fontenc} 

\usepackage[english]{babel} 

\usepackage{amsmath} 

\usepackage[usenames, dvipsnames, svgnames, table]{xcolor} 

\usepackage{tikz} 
\usepackage{tikz-3dplot} 
\usetikzlibrary{arrows.meta}
\usetikzlibrary{shapes,arrows}
\usetikzlibrary{positioning}
\usetikzlibrary{calc}
\usetikzlibrary{shapes.multipart}

\usepackage{pgfplots} 
\pgfplotsset{compat = 1.16}
\usepgfplotslibrary{colorbrewer}

\usepackage[babel]{csquotes} 

\usepackage{float} 

\usepackage{bm} 

\usepackage{amssymb} 

\usepackage{booktabs} 

\usepackage{multirow} 

\usepackage{subcaption} 
\usepackage[inline]{enumitem} 
\setlist{nolistsep} 

\usepackage{hyperref} 
\hypersetup{
    breaklinks = true,
    colorlinks = true,
    bookmarksopen = true,
}
\usepackage[section, above, below]{placeins} 

\usepackage[noabbrev, nameinlink]{cleveref} 

\usepackage[framemethod=tikz]{mdframed} 

\usepackage{makerobust} 

\def\ie{{\textit{i.e.}, }} 
\def\eg{{\textit{e.g.}, }} 

\newcommand{\addcite}[1]{[?]} 

\newcommand{\mysubfigure}[5]{%
  \begin{subfigure}[b]{#1\linewidth}%
    \centering%
    \includegraphics[width=#2\linewidth]{#3}%
    \caption{#4}%
    \label{#5}%
  \end{subfigure}%
}

\newcommand{\revision}[1]{{\textcolor{red}{\textbf{#1}}}}
\renewcommand{\revision}[1]{#1}

\usepackage{regexpatch}
\makeatletter
\regexpatchcmd\ps@pprintTitle 
  {\cE\}\ \cE\}}{\cE\}\cE\}} 
  {}{\FailedToPatch}
\makeatother

%% file: frontmatter.tex
\begin{frontmatter}

\title{Descriptive Modeling of Textiles using\\FE Simulations and Deep Learning}

\journal{Composites Science and Technology}

\author[1,3]{Arturo Mendoza\fnref{fn1}\corref{cor1}}
\cortext[cor1]{Corresponding author.}
\ead{arturo.mendoza-quispe@safrangroup.com}
\author[1]{Roger Trullo\fnref{fn1}}
\author[2]{Yanneck Wielhorski}
\fntext[fn1]{Authors contributed equally.}

\address[1]{Safran Tech, Rue des Jeunes Bois, 78772 Magny les Hameaux, France}
\address[2]{Safran Aircraft Engines, Rond Point R\'en\'e Ravaud-R\'eau, 77550 Moissy-Cramayel, France}
\address[3]{LMT (ENS Paris-Saclay/CNRS/Univ. Paris-Saclay),\\
4 Avenue des Sciences, 91192 Gif-sur-Yvette, France}

\begin{abstract}
\revision{In this work we propose a novel and fully automated method for extracting the yarn geometrical features in woven composites so that a direct parametrization of the textile reinforcement is achieved (\eg FE mesh).
Thus, our aim is not only to perform \emph{yarn segmentation} from tomographic images but rather to provide a complete descriptive modeling of the fabric.
As such, this direct approach improves on previous methods that use voxel-wise masks as intermediate representations followed by re-meshing operations (yarn envelope estimation).}
The proposed approach employs two deep neural network architectures (U-Net and Mask R-CNN).
First, we train the U-Net to generate synthetic CT images from the corresponding FE simulations.
This allows to generate large quantities of \emph{annotated} data without requiring costly manual annotations.
This data is then used to train the Mask R-CNN, which is focused on predicting contour points around each of the yarns in the image.
Experimental results show that our method is accurate and robust for performing yarn instance segmentation on CT images, this is further validated by quantitative and qualitative analyses.
\end{abstract}

\begin{keyword}
\revision{
A. Textile composites
\sep{}
C. Material modelling
\sep{}
Deep learning
\sep{}
D. X-ray computed tomography
\sep{}
C. Finite element analysis (FEA)
}
\end{keyword}

\end{frontmatter}

%% file: 1_introduction/main.tex
Thanks to their elevated specific properties, woven composite materials are widely used in many fields where the final part weight is of upmost importance (\eg aeronautical, aerospace and submarine industries)~\cite{Naresh_MD_2020_review, Mouritz2001ReviewSubmarines}.
These structured composites are obtained by weaving yarns (\eg carbon fiber bundles) in a prescribed pattern.
Given that this weaving pattern is designed so as to achieved certain material properties, a great interest has been given to properly describe the \emph{effective} manufactured arrangement of yarns.
Such is the goal of numerical modeling of textiles: to obtain a geometrical description of the yarns (\ie yarn paths and cross sections) for the final textile configuration.
This can be done using \emph{predictive} approaches that employ numerical simulations, or \emph{descriptive} ones that use (X-ray) tomography images to describe real manufactured parts.

The simplest group of predictive approaches considers the well-established Textile Geometrical Pre-processors (TGP), such as WiseTex~\cite{Lomov_CST_2000, Verpoest_CST_2005} and TexGen~\cite{Sherburn2007_thesis}.
\revision{These approaches model the yarns following only geometrical considerations, hence they constitute a very idealized representation.}
A more advanced group of predictive approaches are those that employ mechanical behavior laws, either at the sub-meso~\cite{Zhou_CST_2004, Mahadik_CompPartA_2010, Durville_IJMF_2010, Green_CS_2014_paper1, Drach_AES_2014, Durville_IJSS_2018} or mesoscale~\cite{Charmetant_CST_2011} (scale of the textile).

On the other hand, descriptive approaches leverage multiple image and data processing techniques to extract useful information from the acquired volume images.
\revision{As such, they can go further than only image segmentation.
Moreover, most of these approaches are based on a local image descriptor called the structure tensor~\cite{Naouar_CS_2014, Straumit_CompPartA_2015, Liu_CS_2017, Wijaya_CompPartA_2019, Wintiba_CS_2020, Naouar_JMS_2020}.}
The local principal image orientations and degree of anisotropy that result from such operation allow to differentiate image voxels belonging to the reinforcement (highly anisotropic) and matrix (highly isotropic).
Hence, it can help distinguishing different yarn orientations (\eg warp and weft) whenever the image contrast and resolution allow it.

\revision{In cases in which these derived features cannot be easily separable (\eg simple threshold on anisotropy level or orientation angle), clustering techniques have had some success.
These approaches aim at finding complex rules that partition the domain spawned by these derived features.}
\revision{In such sense, $k$-means clustering~\cite{Straumit_CompPartA_2015} and Gaussian Mixture Models (GMM)~\cite{Wijaya_CompPartA_2019, Liu_CS_2017, Wintiba_CS_2020} have proven well suited for separating the different yarn orientations (\eg warp and weft yarns).
In the case of 3D woven composites with binder yarns, further work is necessary due to the similarities they share with warp yarns (\eg the same orientation)~\cite{Wintiba_CS_2020}.
It should be noted that, these methods are only capable of distinguishing between yarn orientations but not of isolating each yarn.
For cases in which the yarns are not heavily compacted, elementary morphological operations can be employed~\cite{Naouar_CS_2015}.
However, in more complex cases, such as two yarns of the same type being in contact, some manual intervention may be necessary~\cite{Ewert2020}.
Besides, if an individual identification of the yarns is desired; that is, assign an ID to them, even more steps are required, \eg semi-automatic identification on the first slice and propagation~\cite{Wijaya_CompPartA_2019}.
}

At this point, the output of these methods are no more than binary image masks.
Hence, unless a voxel-based formalism is employed for the material simulation~\cite{Hello_2014_Numerical}, they do necessitate further processing in order to provide usable textile models such as Finite Element (FE) meshes~\cite{Liu_CS_2017, Wintiba_CS_2020, Naouar_CS_2014}.
It is important to note that this category of descriptive methods presents some challenges.
Given that these approaches do not account for spatial information, they often require some clean-up post-processing steps such as morphological operations or smoothing before and after the construction of the textile models.
Indeed, the concatenation of these operations, while useful and effective, can lead to complex pipelines that would seem to demand heavy re-parametrization for every new analyzed sample (\eg re-identification of the mixture model).

Other novel descriptive approaches based on textile models have recently been proposed~\cite{mendoza2019correlation, Benezech_CS_2019}.
In both cases an \textit{a priori} 3D model of the textile is progressively deformed (optimized) so as to conform to the observed sample.
The starting model is obtained by manually extracting the yarn paths and using constant elliptical cross sections.
These latter can progressively be adapted using polygonal shapes until the correct yarn shape is obtained~\cite{Benezech_CS_2019}.
This group of methods (based on \textit{a priori}) do present a great advantage over the previous ones since they directly provide the sought textile model without requiring extra operations.

Finally, some recent descriptive approaches based on deep neural networks have been proposed for performing the \textbf{semantic segmentation} task~\cite{Sinchuk_Materials_2020, Ali_CompPartA_2020_Deep}.
Indeed, given the unprecedented performance of deep learning techniques in other domains (\eg medical imaging, natural images), it seems only natural to apply it into the analysis of materials.
In both cases, \enquote{ready-to-use} network architectures were employed: \citeauthor{Sinchuk_Materials_2020}~\cite{Sinchuk_Materials_2020} used the U-Net architecture~\cite{Ronneberger_2015_UNet} available on the DragonFly software, while \citeauthor{Ali_CompPartA_2020_Deep}~\cite{Ali_CompPartA_2020_Deep} implemented DeepLab v3+~\cite{Chen_2018_DeepLab} in MATLAB using the Deep Learning Toolbox.
Similarly, they both employ a 2D strategy for the segmentation (\ie independently feed slices of the stack to the network), albeit \citeauthor{Ali_CompPartA_2020_Deep}~\cite{Ali_CompPartA_2020_Deep} used a less common top-down approach (the slices are taken in the thickness direction).
To the best of our knowledge, currently, these are the only deep learning-based descriptive methods in the literature.

Moreover, due to the supervised nature of the training that the networks undergo, these approaches require the constitution of a properly annotated database.
This means that for a given tomography slice the corresponding pixel-wise masks are known for each class (\eg warp yarns, weft yarns, matrix).
\citeauthor{Sinchuk_Materials_2020}~\cite{Sinchuk_Materials_2020} employ the \enquote{polygon selections} function in ImageJ while \citeauthor{Ali_CompPartA_2020_Deep}~\cite{Ali_CompPartA_2020_Deep} use the MATLAB Image Labeler Tool.
It should be noted that these semantic segmentation approaches are similar to those based on the structure tensor in the sense that they only segment the yarn orientations (\ie no individualization of yarns) and would require extra post-processing steps to generate usable textile models.

Now, the aim of the method proposed here is to address the yarn segmentation problem as an \textbf{instance segmentation} one using deep learning methods.
Here, \emph{instance} segmentation refers to the process of simultaneously solving the problem of object detection as well as that of \emph{semantic} segmentation.
Additionally, we will achieve such results by treating the problem of segmentation as one of \emph{keypoint estimation}.
Hence, no binary masks of the yarns will be obtained, but rather a set of points describing the outlines of all yarn sections (instances).
To the best of our knowledge, we are the first to propose a method of this nature.
The advantage of this approach is that it provides the yarn cross sections as a collection of control points that can be directly used for constructing the textile model (\eg FE mesh).

Nonetheless, an inherent challenge of supervised deep learning algorithms is the need of considerable amounts of annotated datasets during training.
This crucial information is very expensive to obtain since considerable manual operations are required for properly annotating all yarns in an image (\ie delimiting their cross sections and paths), even when some semi-automated procedures are employed~\cite{Ali_CompPartA_2020_Deep}.
Here, we propose to circumvent this problem by using a limited amount of (imprecise) annotations in a \emph{weakly supervised} manner rather than a fully supervised one.
In particular, we will employ a deep neural network for creating synthetic CT images from these weakly annotated images, we refer to these as pseudo-CT images.
After training, the network will be used on \enquote{annotated} images that come from FE simulations (the yarn information is numerically accessible).
As such, we are now capable of generating many pseudo-CT images where the annotations come from simulations rather than by manual operations.
Finally, another deep network will take this completely numerical database for training the instance segmentation model, which will later be used on real images.

The available materials and an overview of the method will given in~\cref{sec:materials_and_method}.
The generation of pseudo-CT images and the instance segmentation steps will be detailed in~\cref{sec:generation_pseudo,sec:instance_segmentation} respectively.
The use of these results for obtaining textile models will be discussed in~\cref{sec:mesh_generation}.
Finally, a quantitative evaluation of the obtained results will be proposed in~\cref{sec:validation}.

%% file: 2_methods/main.tex
\input{1_materials}
\input{2_method}

%% file: 2_methods/1_materials.tex
\subsection{Tomographic volume}
\label{sec:tomographic_volume}

The 3D woven fabric studied in this work is a ply-to-ply angle-interlock, composed of 75 carbon fiber yarns: 39 warp and 36 weft.
\revision{All yarns are of the same type and size.}
The warp yarns are distributed alternately in a sequence of 4 and 3 yarns in two consecutive columns. 
Similarly, two consecutive weft columns are composed of 5 and 4 yarns respectively.
In this sample, there are 11 warp planes and 8 weft columns.

The sample was scanned with a GE Phoenix-Xray tomograph (GE v|tome|x L300) at a resolution of 20$\mu m$.
The reconstructed image $I_{CT}$ is $1698 \times 1814 \times 402$ voxels in size, and is normalized so that its dynamic range is [0, 1].
In the following, the X, Y and Z axes are aligned with the warp, the weft and the thickness orientations respectively.
This information is also given in~\cref{table:summarize_image}, which summarizes all relevant information for the employed images.

\subsection{Simple textile model}
\label{subsec:simple_textile_model}

The creation of a geometrical textile model from the acquired CT volume $I_{CT}$ was done in a semi-automatic manner.
\revision{First, the yarn paths are manually identified by locating the coordinates of their cross-over points where crimp is maximum ; that is, at intersections of warp and weft directions.
These are defined for each column in the orthogonal direction (\eg 8 cross-over points for the warp and 11 for the weft) and are manually placed at the approximate center of the corresponding cross section.}
Then, a B-spline interpolation is used with this first set of control points, and if required, each yarn path can be further adjusted by adding more control points.
As such, the final number of control points is not necessary identical for each yarn.
These local adjustments are rather necessary in zones with high crimp.

No identification of the yarn cross section is performed for this model.
\revision{Hence a uniform elliptical shape is attributed to all 75 yarns, uniform all along each yarn and equal for all yarns.
A first estimation of the ellipse radii is obtained by manually measuring some yarn sections in a collection of carefully selected CT volume images.
Since in practice, the intra-yarn fiber volume fraction is not constant all along the yarn path, the radii are then adjusted to better match the theoretical overall fiber volume fraction value of the composite.
}

\subsection{Mechanical simulations}
\label{sec:simulations}

Numerical Finite Element (FE) calculations are carried out using the in-house software Multifil~\cite{Durville_IJSS_2018} devoted to simulating mechanical behaviors of entangled fibrous materials using a quasi-static formulation.
It is based on the solution of the mechanical equilibrium of a collection of beams in contact-friction interaction.
\revision{Here, each yarn is modeled using 61 beams (also called virtual fibers or macro-filaments) twisted together so as to mimic the internal cohesion of fibers.
Given that the employed number of virtual fibers is sufficient for properly describing the deformation of the yarn cross section~\cite{Daelemans_CST_2021}, each beam is described by the isotropic Saint-Venant law.}

\revision{This algorithm proposes a novel strategy for obtaining the \enquote{as-woven} state.
Instead of starting from a nominal configuration close to the \enquote{as-woven} state (as most methods do~\cite{Stig_CS_2012_paper1, Green_CS_2014_paper1}), the starting configuration places all yarns in the same plane~\cite{Durville_IJMF_2010, Durville_IJSS_2018}.}
Hence, the algorithm progressively separates the virtual fibers using a contact-friction algorithm until reaching the global mechanical equilibrium of the targeted weaving pattern.

\revision{The goal of the mechanical simulations presented here is to provide a rich and useful information for the subsequent deep learning models.
As such, instead of just using the textile model at the \enquote{as-woven} state, we propose to perform further compaction simulations on this model.
These latter will provide a much greater variability of textile density.}

\revision{The weaving pattern used for the textile model is the same as in the acquired CT volume $I_{CT}$.
However, we only considered 64 yarns for the simulation: 28 warp and 36 weft (\ie 11 warp less than in the original volume).
Once the as-woven configuration is reached (mechanical equilibrium), a subsequent sub-mesoscale simulation of transverse compression is performed until a desired final compacted state is achieved.}
Assuming an initial thickness $H_{init}$ and a final $H_{final}$, the increment compression is $(H_{init}- H_{final}) / N_{steps}$ with $N_{steps} = 12$.
These different compaction steps emphasize the variability of the textile morphology during a forming process (\eg changes in both yarn trajectories and cross sections).
Note that the simulation step closest to that of the original sample is the initial one (\ie the as-woven state).

All the 12 simulation steps are considered so as to obtain their respective textile models.
However, given that these simulations (determination of the initial configuration and transverse compression) are performed at the sub-mesoscale, the yarn envelopes need to be reconstructed from the virtual fibers.
\revision{This considers the determination of the yarn paths and the estimation of the corresponding yarn cross sections.}

Here, the yarn cross sections were computed at 38 and 49 locations along the warp and weft yarn paths.
The, 100 points were used for estimating the polygonal envelopes for all cross sections, but only 10 equally sampled points are taken for the following.
Additionally, the center point (the centroid of all cross section points) is considered for each cross section.
Hence, 11 points are used for representing the yarn at each location.

\subsection{Voxel conversion}
\label{sec:voxel_conversion}

The previous simple textile model and the 12 textile models from FE simulations are expressed using a parametric formalism.
In order to convert them into \enquote{images}, the in-house software REVoxel~\cite{Hello_2014_Numerical} is used for the voxel conversion.
This software is capable of handling this type of parametric description since it was developed for predicting the mechanical properties of woven composite materials using voxel-FE models.

The produced volumes are called $I_{GT}$ (as in ground truth) for the simple textile model and $I_{FE}^{(k)}$ with $k \in [1, 12]$ for the numerically obtained models.
The assigned intensity levels (\ie gray values) are the yarn IDs, hence the dynamic range is as large as the number of yarns (75 and 64, respectively).
Moreover, these \emph{labeled} images are calculated at the same resolution (\ie voxel size) as that of the acquired CT volume $I_{CT}$.
Thus, $I_{GT}$ and $I_{CT}$ have the same size, but $I_{FE}^{(k)}$ is $1401 \times 1401 \times 321$ voxels in size.
This smaller thickness (321 voxels) results from the average thickness from all twelve simulation steps and is positioned so that half-plane (orthogonal to the thickness) remains constant.
Then, in the generated images, the compaction seems to be symmetric with respect to a center plane.

\subsection{Extraction of 2D slices}

In the following, all 3D volumes will be analyzed through 2D slices taken along the warp and weft orientations.
As such, 3512 slices (1814 XZ slices and 1698 YZ slices) can be extracted from both images $I_{CT}$ and $I_{GT}$.
Whereas only 2802 slices (1401 XZ slices and 1401 YZ slices) can be extracted from each of the twelve images $I_{FE}^{(k)}$.
These collections of 2D images are gathered into the datasets $x_{CT}[i]$, $x_{GT}[i]$ and $x_{FE}^{(k)}[j]$ for $i \in [1, 3512]$ and $j \in [1, 2802]$.

\revision{This approach, known as 2.5D in the medical field, is extremely advantageous because it will allow to use existing architectures and frameworks, most of which are developed for dealing with 2D images rather than 3D volumes.}

\begin{table}
    \centering
    \caption{
    Summary of available images and datasets}
    \begin{tabular}{@{}cccc@{}}
        \toprule
        Name & Notation & Number of yarns & Image size \\
        \midrule
        Acquired CT         & $I_{CT}$  & 75 (39 warp) & $1698 \times 1814 \times 402$ \\
        Simple label        & $I_{GT}$  & 75 (39 warp) & $1698 \times 1814 \times 402$ \\
        FE \enquote{labels} & $I_{FE}$  & 64 (28 warp) & $1401 \times 1401 \times 321$ \\
        Pseudo-CT           & $I_{pCT}$ & 64 (28 warp) & $1401 \times 1401 \times 321$ \\
        \bottomrule
    \end{tabular}
    \label{table:summarize_image}
\end{table}

%% file: 2_methods/2_method.tex
\subsection{Overview of the method}

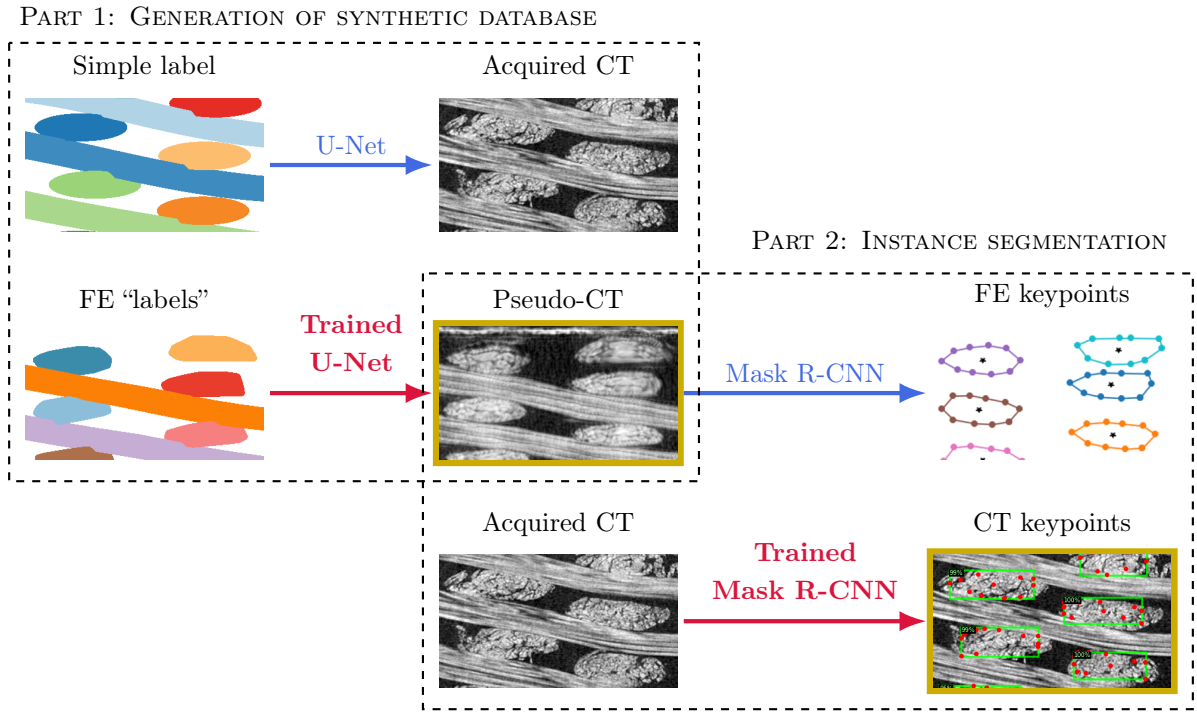
\begin{figure}[hb]
    \centering
    \begin{tikzpicture}[
        auto, thick, node distance=0.1em,
        border/.style={anchor=south west, inner sep=0, draw, line width=0.5em},
        arrow/.style={ultra thick, -Latex},
        label/.style={}]
        \newcommand{\nodefigure}[1]{\includegraphics[width=0.19\linewidth]{figures/#1.png}}
        %
        \node [border, white] at (0,0) (x1) {\nodefigure{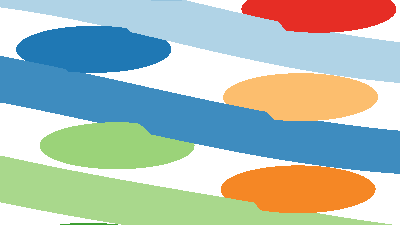}};
        \node [border, white, right =of x1, xshift=6em] (x2) {\nodefigure{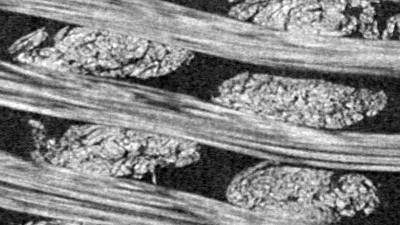}};
        \node [border, white, below =of x1, yshift=-3.0em] (x3) {\nodefigure{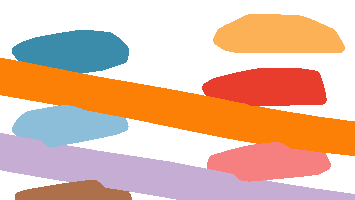}};
        \node [border, Gold!80!Black, right =of x3, xshift=6em] (x4) {\nodefigure{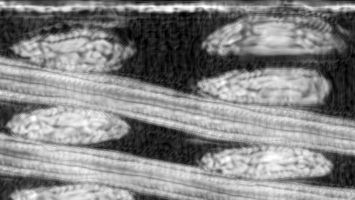}};
        \node [border, white, right =of x4, xshift=9em] (x5) {\nodefigure{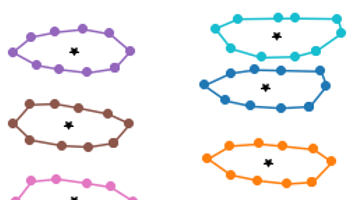}};
        \node [border, white, below =of x4, yshift=-3.0em] (x6) {\nodefigure{unet_tomo}};
        \node [border, Gold!80!Black, right =of x6, xshift=9em] (x7) {\nodefigure{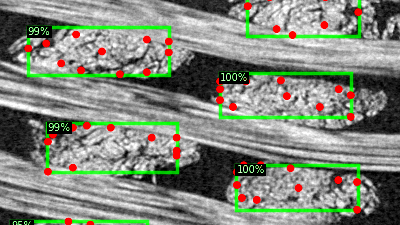}};
        %
        \node [label, above =of x1] (x1p) {Simple label};
        \node [label, above =of x2] (x2p) {Acquired CT};
        \node [label, above =of x3] (x3p) {FE \enquote{labels}};
        \node [label, above =of x4] (x4p) {Pseudo-CT};
        \node [label, above =of x5] (x5p) {FE keypoints};
        \node [label, above =of x6] (x6p) {Acquired CT};
        \node [label, above =of x7] (x7p) {CT keypoints};
        %
        \draw[arrow, RoyalBlue] (x1) -- node [auto] {U-Net} (x2);
        \draw[arrow, Crimson ] (x3) -- node [auto] {\begin{tabular}{c}\textbf{Trained} \\ \textbf{U-Net}\end{tabular}} (x4);
        \draw[arrow, RoyalBlue] (x4) -- node [auto] {Mask R-CNN} (x5);
        \draw[arrow, Crimson ] (x6) -- node [auto] {\begin{tabular}{c}\textbf{Trained} \\ \textbf{Mask R-CNN}\end{tabular}} (x7);
        %
        \node at (x1p -| x1.west) (x1o) {};
        \node at (x4p -| x4.west) (x4o) {};
        \draw[dashed] ($(x1o.north west)+(0,0.2)$) rectangle ($(x4.south east)+(0.2,-0.2)$);
        \draw[dashed] ($(x4o.north west)+(0,0.2)$) rectangle ($(x7.south east)+(0.2,-0.2)$);
        \node at ($(x1o.north west)+(0,0.3)$) [above right] {\textsc{Part 1: Generation of synthetic database}};
        \node at ($(x5p.north east)+(0.5,0.2)$) [above left] {\textsc{Part 2: Instance segmentation}};

    \end{tikzpicture}
    \caption{\revision{General overview of the method showing the training and inference phases of both architectures}, both main parts of the method are delimited, and the results of each part are bordered in \textcolor{Gold!80!Black}{\textbf{gold}}.}
    \label{fig:summary_paper}
\end{figure}

As it can be seen in the diagram of the proposed method in~\cref{fig:summary_paper}, it is composed of two main parts:
\begin{enumerate*}[label=(\roman*), font=\itshape]
    \item the generation of a virtual dataset using a simple annotations, and
    \item the use of this dataset for training an instance segmentation model.
\end{enumerate*}
In both cases, deep neural networks are employed.

The first part, detailed in~\cref{sec:generation_pseudo}, consists in \revision{performing a sort of \enquote{inverse segmentation}} that allows converting labeled images into realistic CT images (\ie pseudo-CT).
Here, the \enquote{simple labels} obtained from the simple textile model are used as input during the training phase for a U-Net network~\cite{Ronneberger_2015_UNet}.
Then, the images obtained from mechanical simulations can be used as inputs for the network (during inference) so as to obtain the corresponding pseudo-CT images, which constitute the database for the next part.

The second part, detailed in~\cref{sec:instance_segmentation}, employs a Mask R-CNN architecture~\cite{mask_rcnn} for performing the instance segmentation task.
Here, the \enquote{perfectly} annotated images that come from the mechanical simulations and their corresponding pseudo-CT images constitute an entirely numerical source of information for the network.
Indeed, the network only encounters \emph{real} CT images when used during inference, never during training.
Then, this network can be used on the acquired CT image for obtaining the description of the yarn cross sections using a set of keypoints.

%% file: 3_unet/main.tex
\subsection{Pre-processing of labeled images}
\input{1_pre_processing}

\subsection{Neural network architecture}
\input{2_architecture}

\subsection{Training of the network}
\input{3_training}

\subsection{Implementation details}
\input{4_implementation}

\subsection{Results}
\input{5_results}

%% file: 3_unet/1_pre_processing.tex
The labeled images, obtained after voxelization of either the real CT or the FE simulations, contain voxels whose values correspond to the yarn ID.
This information is not useful for the creation of the sought pseudo-CT images.
Hence, they will be given new labels that represent the yarn type.
Two strategies are possible:
\begin{enumerate*}[label=(\roman*), font=\itshape]
    \item distinguish between warp and weft orientations, or
    \item distinguish between yarns seen longitudinally and yarns seen transversally.
\end{enumerate*}

The first approach presents the advantage of \enquote{encoding} the yarn type in the labels.
However, given the 2D nature of the analysis, any given yarn (\eg a warp yarn) will sometimes be seen longitudinally (in XZ slices) and sometimes transversally (in YZ slices).
This is problematic for the network since it needs to comprehend this duality.
Indeed, our initial experiments using this approach failed to provide satisfactory results since it only seemed to work for one yarn direction but not the other.

On the other hand, the second approach has the advantage of always presenting the transversally seen yarns using the same label, independently of the yarn type (\ie warp or weft).
\revision{Also, even if the yarn type is \enquote{lost} in the new labels, it can still be deduced from the images.
Indeed, as it can be seen in~\cref{sfig:tomo_warp_tomo_0,sfig:tomo_weft_tomo_0} the warp yarns (seen longitudinally) present a much higher wave amplitude than the weft yarns.
This type of contextual information should be easily captured by a convolutional neural network.}
For this reason, the second approach is taken hereinafter.

\revision{The voxel-wise transformation function $f_{tri}$ is constructed so that for any 2D input image (\eg $x_{GT}[i]$ for $i \in [1, 3512]$) with various values per voxel is converted into a \enquote{trinary} image ${f_{tri}(x_{GT}[i])}$ where the gray levels 1 (white), 0.5 (gray) and 0 (black) are assigned for the visible yarn cross sections, the longitudinally visible yarns, and the air (or matrix constituent) respectively.
For example, see the conversion from~\cref{sfig:tomo_warp_label_0} to~\cref{sfig:tomo_warp_trinary_0} where the warp yarns are labeled 1 and the weft yarns 0.5.
Similarly, in~\cref{sfig:tomo_weft_label_0} to~\cref{sfig:tomo_weft_trinary_0} with 0.5 for warp yarns and 1 for weft yarns.}

%% file: 3_unet/2_architecture.tex
The goal is to construct a mapping function $f_{pseudo}$ so that
\begin{equation}
    x_{CT}[i] \approx f_{pseudo}(x_{GT}[i]) = f_{unet}^{\prime} \circ f_{tri}(x_{GT}[i])
\end{equation}
with $f_{unet}^{\prime}$ a slightly modified version of the U-Net network~\cite{Ronneberger_2015_UNet}.

The U-Net architecture consists of a contracting part to capture context and a symmetric expanding part that enables precise localization.
The contracting part follows the architecture of a typical convolutional network.
It consists of repeated application of convolutions, each followed by a rectified linear unit (ReLU) and a max pooling operation.
Next, the expanding part is composed of a sequence of up-convolutions and concatenations with high-resolution features from the contracting part (via skip-connections).
This u-shaped architecture combines high-resolution (spatial) information from the contracting part and the condensed feature information that is successively up-sampled.

\begin{figure}
    \centering
    \begin{tikzpicture}[auto, thick, node distance = 1em, arrow/.style={-Latex}]
        
        \node [] at (0,0) [] (x0) {$u$};
        \node [below =of x0] (x0a) {};
        \node [below =of x0a] (x0b) {};
        \node [right =of x0b] (x0c) {};

        \node [draw, right =of x0c] (x1) {$\texttt{Conv2}^{1}_{64}$};
        \node [draw, below =of x1] (x2) {$\texttt{Down}^{64}_{128}$};
        \node [draw, below =of x2] (x3) {$\texttt{Down}^{128}_{256}$};
        \node [draw, below =of x3] (x4) {$\texttt{Down}^{256}_{512}$};
        \node [draw, below right =of x4] (x5) {$\texttt{Down}^{512}_{512}$};

        \node [draw, above right =of x5] (x6) {$\texttt{Up}^{1024}_{256}$};
        \node [draw, above =of x6] (x7) {$\texttt{Up}^{512}_{128}$};
        \node [draw, above =of x7] (x8) {$\texttt{Up}^{256}_{64}$};
        \node [draw, above =of x8] (x9) {$\texttt{Up}^{128}_{64}$};
        \node [draw, right =of x9] (x10) {$\texttt{Conv}^{64}_{1}$};

        \draw[thick, arrow] (x0) -- +(0.45,0) |- (x1);
        \draw[arrow] (x1) -- (x2);
        \draw[arrow] (x2) -- (x3);
        \draw[arrow] (x3) -- (x4);
        \draw[arrow] (x4) |- (x5);

        \draw[arrow] (x5) -| (x6);
        \draw[arrow] (x6) -- (x7);
        \draw[arrow] (x7) -- (x8);
        \draw[arrow] (x8) -- (x9);
        \draw[arrow] (x9) -- (x10);

        \draw[arrow] (x4) -- (x6);
        \draw[arrow] (x3) -- (x7);
        \draw[arrow] (x2) -- (x8);
        \draw[arrow] (x1) -- (x9);

        \draw[arrow] (x9) -- (x10);

        \node [right =of x10] (x10b) {};
        
        \node[circle, draw, inner sep=0.4em, label=center:{$+$}] at (x0 -| x10b) (x11) {};
        \draw[] (x0) -- (x11);

        \node [draw, right =of x11] (x12) {\texttt{Sigm}};
        \draw[arrow] (x11) -- (x12);

        \node [right =of x12] (x13) {$f_{unet}^{\prime}(u)$};
        \draw[arrow] (x12) -- (x13);

        \node [gray] at (x10 -| x13) (x14) {$f_{unet}(u)$};
        \draw[gray, arrow] (x10) -- (x14);
        \draw[] (x10.east) -| (x11.south); 

        \node [] at (x5 -| x10) (aux_x5_x10) {};
        \draw[dashed] ($(x1.north west)+(-0.35,0.2)$) rectangle ($(aux_x5_x10.south east)+(0.80,-0.35)$);
        \node [] at ($(x1.north west)+(-0.35,0.2)$) [above right] {U-Net};

    \end{tikzpicture}
    \caption{Diagram of the modified U-Net architecture~\cite{Ronneberger_2015_UNet}}
    \label{fig:unet_architecture}
\end{figure}
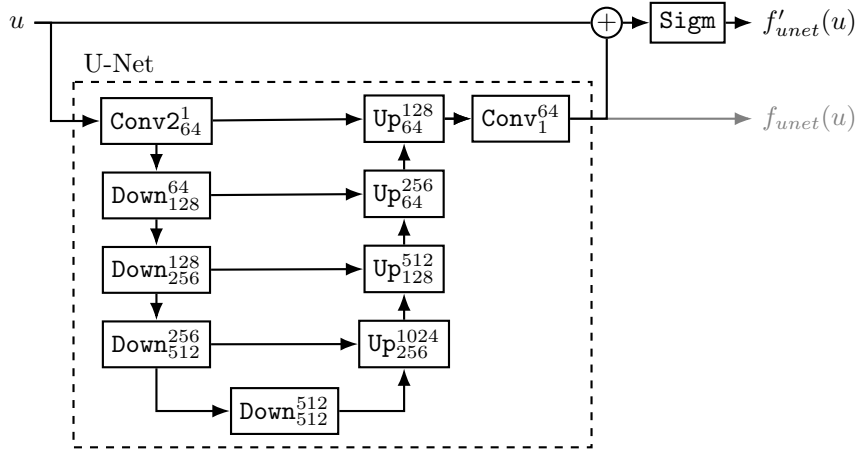

The modification proposed here is two-fold.
First, a long skip-connection is added between the input and output of the last convolution layer.
\revision{This type of connection has proven to be useful in image generation tasks without affecting computational times~\cite{super_res}}.
Second, a sigmoid activation function is added to this new output so that the function is bounded to the range [0, 1].
This can be written as
\begin{equation}
    f_{unet}^{\prime}(u) = \texttt{Sigm}(f_{unet}(u) + u)
\end{equation}
for any input image $u$, with $f_{unet}$ the \enquote{classical} U-Net network~\cite{Ronneberger_2015_UNet} and $\texttt{Sigm}$ the sigmoid function.
\Cref{fig:unet_architecture} schematizes the U-Net architecture.
For completeness, the following functions are employed:
\begin{align}
    \texttt{Conv1}^{a}_{b}(u) &= \texttt{ReLU} \circ \texttt{BatchNorm} \circ \texttt{Conv}^{a}_{b} (u) \\
    \texttt{Conv2}^{a}_{b}(u) &= \texttt{Conv1} \circ \texttt{Conv1} (u)  \\
    \texttt{Conv3}^{a}_{b}(u) &= \texttt{Conv2} \circ \texttt{Conv1} (u)  \\
    \texttt{Down}^{a}_{b}(u) &= \texttt{Conv2}^{a}_{b} \circ \texttt{MaxPool} (u) \\
    \texttt{Up}^{a}_{b}(u, v) &= \texttt{Conv2}^{a}_{b} \circ \texttt{Concat} (\texttt{Upsample}(u), v)
\end{align}
with $a$ and $b$ denoting the sizes of input and output feature spaces.
For further information or more details for each of these, along with the \texttt{ReLU}, \texttt{Sigm}, \texttt{BatchNorm}, \texttt{MaxPool}, \texttt{Concat}, \texttt{Upsample}, \texttt{Dense} and \texttt{Conv} operations, see the respective references~\cite{Ronneberger_2015_UNet,Simonyan_2015_VGG}.

%% file: 3_unet/3_training.tex
The network is trained from scratch on a random 80\% of the $x_{GT}$ and $x_{CT}$ datasets (2810 image pairs).
The remaining 20\% (702 image pairs) is used for validation purposes.
No test set is needed here since the network will be used on the virtual images issued from simulation.

The optimization is performed with respect to a custom loss function
\begin{equation}
    \mathcal{L}_{unet}(u, w) = \mathcal{L}_{pix}(u, w) + \mathcal{L}_{per}(u, w)
\end{equation}
for any input image $u$ and any target image $w$, with
\begin{equation}
    \mathcal{L}_{pix}(u, w) = MSE(u, w)
\end{equation}
a classical pixel-wise $L_{2}$ metric, and $\mathcal{L}_{per}$ a perceptual loss~\cite{Johnson_2016_Perceptual} that encourages the output image to be \enquote{perceptually similar} to the target defined as
\begin{equation}
    \mathcal{L}_{per}(u, w) = MSE(f_{vgg}^{\prime}(u), f_{vgg}^{\prime}(w))
\end{equation}
The latter, also called feature reconstruction loss, aims to measure high-level perceptual and semantic differences between images~\cite{Johnson_2016_Perceptual}.
It makes use of a \emph{loss network} $f_{vgg}^{\prime}$ pre-trained for image classification, meaning that $\mathcal{L}_{per}$ is itself a deep convolutional neural network.
Here, $f_{vgg}^{\prime}$ is a subset of the 16-layer VGG network (\eg VGG-16)~\cite{Simonyan_2015_VGG} pre-trained on a subset of the ImageNet dataset~\cite{Russakovsky_2015_ImageNet}, a large database with more than one million of manually annotated natural images.

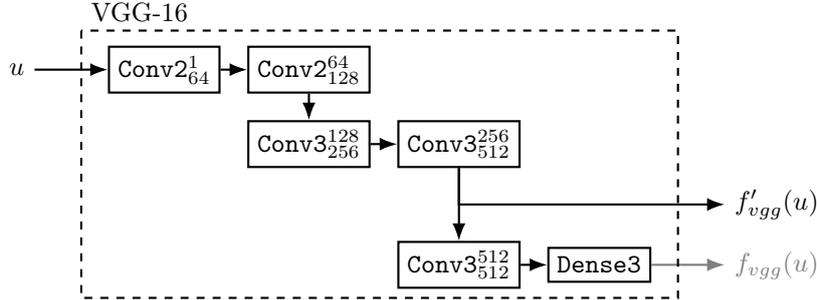
\begin{figure}
    \centering
    \begin{tikzpicture}[auto, thick, node distance = 1em, arrow/.style={-Latex}]

        \node [] at (0,0) [] (y0) {$u$};

        \node [] [right =of y0] (y0b) {};

        \node [draw, right =of y0b] (y1) {$\texttt{Conv2}^{1}_{64}$};
        \node [draw, right =of y1] (y2) {$\texttt{Conv2}^{64}_{128}$};
        \node [draw, below =of y2] (y3) {$\texttt{Conv3}^{128}_{256}$};
        \node [draw, right =of y3] (y4) {$\texttt{Conv3}^{256}_{512}$};
        \node [below =of y4] (y4b) {};
        \node [draw, below =of y4b] (y5) {$\texttt{Conv3}^{512}_{512}$};
        \node [draw, right =of y5] (y6) {$\texttt{Dense3}$};

        \draw[thick, arrow] (y0) -- +(0.45,0) |- (y1);
        \draw[arrow] (y1) -- (y2);
        \draw[arrow] (y2) -- (y3);
        \draw[arrow] (y3) -- (y4);
        \draw[arrow] (y4) -- (y5);
        \draw[arrow] (y5) -- (y6);

        \node [right =of y6] (y6b) {};
        \node [gray, right =of y6b] (y7) {$f_{vgg}(u)$};
        \draw[gray, arrow] (y6) -- (y7);

        \node [] at (y4b -| y7) (y8) {$f_{vgg}^{\prime}(u)$};
        \draw[arrow] (y4) |- (y8);

        \draw[dashed] ($(y1.north west)+(-0.35,0.2)$) rectangle ($(y6.south east)+(0.35,-0.2)$);
        \node [] at ($(y1.north west)+(-0.35,0.2)$) [above right] {VGG-16};

    \end{tikzpicture}
    \caption{Diagram of the standard VGG-16~\cite{Simonyan_2015_VGG} architecture}
    \label{fig:vgg_architecture}
\end{figure}

Selecting a subset of the complete network allows the evaluation of reconstructed features to be performed at an \enquote{inner} layer.
Indeed, while using early layers will tend to favor visually indistinguishable images, later layers will favor image content and overall spatial structure over color, texture, and exact shape.
Here, the chosen layer is last (the third) of the fourth block (out of five in total).
This desired \texttt{relu4\_3}~\cite{Johnson_2016_Perceptual, Simonyan_2015_VGG} output corresponds to the function $f_{vgg}^{\prime}$, while $f_{vgg}$ corresponds to the output of the entire VGG-16 architecture, see~\cref{fig:vgg_architecture} for details on the location of these functions.

\revision{In this context, the VGG-16 is being used as a \enquote{backbone} or feature extracting network.
It should be noted that the choice of the network itself or the dataset it was trained on are relatively arbitrary.
Yet, the sheer amount of annotated images in the ImageNet dataset and (pre-)training of the VGG-16 network guarantee a very general feature extractor useful for describing any type of images (from ImageNet or otherwise).
}

Finally, the training seeks to minimize $\mathcal{L}_{unet}(x_{GT}, x_{CT})$ for the labeled images $x_{GT}$ as inputs and the CT images $x_{CT}$ as output.
\revision{Note that the weights of the VGG-16 network do not need updating during training.}

%% file: 3_unet/4_implementation.tex
The experiment was carried out using the library PyTorch Lightning~\cite{falcon_2019_lightning} since it provides a high-level interface to PyTorch~\cite{paswke_2019_pytorch}.
Both open-source Python libraries facilitate the access to deep learning experiments.
In fact, the framework provided by PyTorch Lightning organizes PyTorch code so as to dissociate research from engineering.
In practice this means that scalable deep learning models can easily be created and run on distributed setups while keeping the models almost intact (\ie hardware agnostic).

Hence, once the models were tested at a small scale, they were easily deployed on 6 GPU Nvidia Tesla T4 (16Gb RAM) using the \enquote{Distributed Data Parallel} accelerator on PyTorch Lightning.
The gradient descent algorithm used is Adam~\cite{kingma_2015_adam}.
Moreover, the defaults proposed by PyTorch and PyTorch Lightning were used for all other parameters (\eg cross-validation, event logging) and hyperparameters (\eg learning rate, batch size).

Yet, special attention was required for the batches.
As previously explained, the dataset is composed by the collection of XZ and YZ slices from both volumes.
However, these slices have different sizes, $1698 \times 402$ and $1814 \times 402$ respectively.
Then, whenever a YZ slice is (randomly) chosen to form a batch, it is horizontally trimmed so that they all have the same size of $1698 \times 402$ pixels.

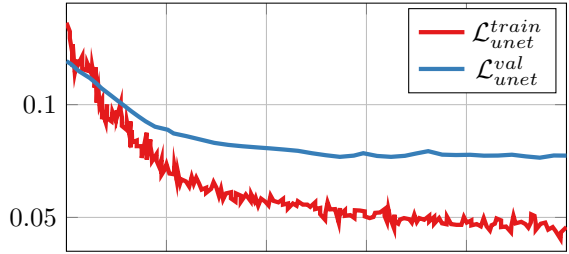
\begin{figure}
    \centering
    \begin{tikzpicture}
        \begin{axis}[
            scale only axis=true, cycle list/Set1,
            width=0.4\linewidth, height=0.2\linewidth,
            xmin=0, xmax=1, ymin=0.035, ymax=0.145,
            grid=both, xticklabels={,,}, scaled y ticks=false,
            yticklabel style={/pgf/number format/fixed,/pgf/number format/precision=5},
            ]
            \addplot+ [ultra thick, mark=none] table [x index=0, y index=1, col sep=comma] {3_unet/figures/loss/training.csv};
            \addplot+ [ultra thick, mark=none] table [x index=0, y index=1, col sep=comma] {3_unet/figures/loss/validation.csv};
            \addlegendentry{$\mathcal{L}_{unet}^{train}$};
            \addlegendentry{$\mathcal{L}_{unet}^{val}$};
        \end{axis}
    \end{tikzpicture}
    \caption{Loss evolution for training and validation datasets}
    \label{fig:unet_loss_evolution}
\end{figure}

The loss evolution for the training and testing datasets is shown in~\cref{fig:unet_loss_evolution}.
Training was stopped after 40 epochs since it became clear that the \enquote{total} loss function on the validation dataset $\mathcal{L}_{unet}^{val}$ had attained a plateau.
This empirical early-stop was performed so as to avoid over-fitting.

%% file: 3_unet/5_results.tex
\begin{figure}
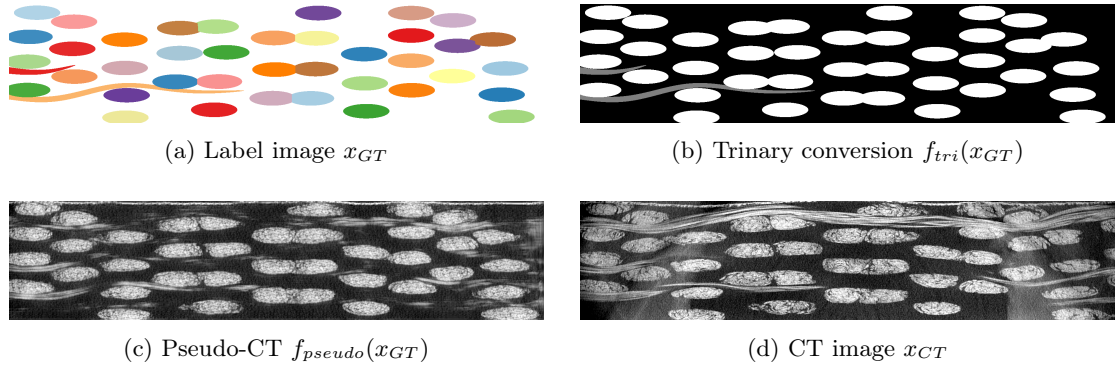

    \centering
    \mysubfigure{0.45}{0.95}{figures/tomo/warp_label_0.png}{Label image $x_{GT}$}{sfig:tomo_warp_label_0}
    \mysubfigure{0.45}{0.95}{figures/tomo/warp_trinary_0.png}{Trinary conversion $f_{tri}(x_{GT})$}{sfig:tomo_warp_trinary_0}
    \par\bigskip

    \mysubfigure{0.45}{0.95}{figures/tomo/warp_pseudo_0.png}{Pseudo-CT $f_{pseudo}(x_{GT})$}{sfig:tomo_warp_pseudo_0}
    \mysubfigure{0.45}{0.95}{figures/tomo/warp_tomo_0.png}{CT image $x_{CT}$}{sfig:tomo_warp_tomo_0}
    \caption{Initial YZ slices on the CT and label volumes, the weft yarns are seen longitudinally, reconstruction artifacts are visible in the CT image}%
    \label{fig:unet_tomo_warp_artifacts}
\end{figure}
\begin{figure}
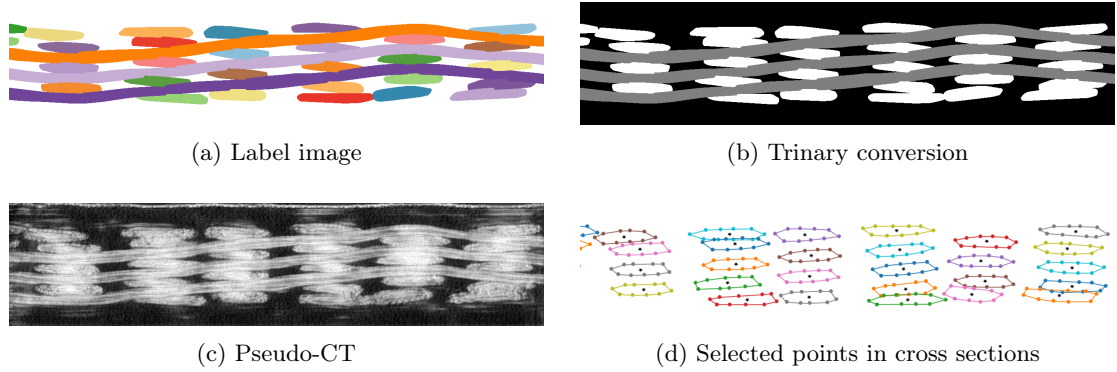

    \centering
    \mysubfigure{0.45}{0.95}{figures/fe/weft_label_100.png}{Label image}{sfig:fe_weft_label_100}
    \mysubfigure{0.45}{0.95}{figures/fe/weft_trinary_100.png}{Trinary conversion}{sfig:fe_weft_trinary_100}
    \par\bigskip

    \mysubfigure{0.45}{0.95}{figures/fe/weft_pseudo_100.png}{Pseudo-CT}{sfig:fe_weft_pseudo_100}
    \mysubfigure{0.45}{0.95}{figures/fe/weft_points_100.png}{Selected points in cross sections}{sfig:fe_weft_points_100}
    \caption{Mid-XZ slices for the last simulation step, the warp yarns are seen longitudinally}%
    \label{fig:unet_fe_weft_2}
\end{figure}

Once the network has been trained, the VGG-16 backbone is no longer needed for inference, only the modified U-Net.
As such, the model can be used for predicting on the already seen labeled images for the CT volume $I_{GT}$ (see~\cref{fig:unet_tomo_warp_artifacts,fig:unet_tomo_weft_artifacts}), as well as on the twelve simulated volumes $I_{FE}$ (see~\cref{fig:unet_fe_warp_1,fig:unet_fe_weft_1,fig:unet_fe_warp_2,fig:unet_fe_weft_2}).
Moreover, the network can also be used on \enquote{hand-drawn} images representing other textile architectures as it can be seen in~\ref{appendix:hand_drawn}.
A visualization of the manner in which the network progressively obtains these results is given in~\ref{appendix:features}.

The performance displayed in~\cref{fig:unet_tomo_warp_artifacts,fig:unet_tomo_weft_artifacts} is of very high quality.
The model is capable of generating extremely realistic pseudo-CT images, comparable to the real one.
Moreover, the generated pseudo-CT images do not present any tomographic reconstruction artifacts (\eg ring and cupping artifacts), unlike the original real ones.

It is important to note that the model learnt to transform the simple elliptical cross sections into more complex ones just by observing the CT volume, which clearly does not present yarns with perfect ellipses.
Also, it is capable of better handling the proximal regions between yarns.
As it can be seen, the in-house software used for voxel conversion~\cite{Hello_2014_Numerical} at times creates some undesirable \enquote{lips} in these regions (see~\cref{sfig:tomo_warp_label_0,sfig:tomo_weft_label_0}).
Yet, the model has properly learnt to not reproduce this numerical artifact.
In conclusion, the model is capable of overcoming both numerical limitations thanks to the learning process.

The results for the simulated textile, shown in~\cref{fig:unet_fe_warp_1,fig:unet_fe_weft_1,fig:unet_fe_warp_2,fig:unet_fe_weft_2} demonstrate that creating realistic CT images from completely numerical results is possible.
The quality of the pseudo-CT images produced for a median compaction level (6th simulation step), shown in~\cref{fig:unet_fe_warp_1,fig:unet_fe_weft_1} is so convincing that, at a first glance, one could be fooled into taking them as real CT images.
This is less true for the highest compaction level (12th simulation step), shown in~\cref{fig:unet_fe_warp_2,fig:unet_fe_weft_2}, where the inner yarn texture is less faithful to such a case.
However, it is important to remark that real samples displaying such extreme compression have not been seen by the network during training.
Hence, this case requires some careful extrapolation from the network.

It is also interesting to note that the case of the simulated textile presents a relevant challenge.
As it was stated, the network has learnt to convert simple elliptical cross sections into realistic ones.
This \enquote{problem} is non-existent for the images issued from our simulation, since the employed software (MultiFil) produces \enquote{correct} yarn cross sections.
Interestingly, the model mostly \emph{respects} the input yarn cross sections, but in some cases it deforms them further.
This process is guided by some global spatial clues that the network had to learn for the images in the training dataset.

\begin{figure}
    \centering
    \mysubfigure{0.45}{0.95}{figures/tomo/weft_label_0.png}{Label image $x_{GT}$}{sfig:tomo_weft_label_0}
    \mysubfigure{0.45}{0.95}{figures/tomo/weft_trinary_0.png}{Trinary conversion $f_{tri}(x_{GT})$}{sfig:tomo_weft_trinary_0}
    \par\bigskip

    \mysubfigure{0.45}{0.95}{figures/tomo/weft_pseudo_0.png}{Pseudo-CT $f_{pseudo}(x_{GT})$}{sfig:tomo_weft_pseudo_0}
    \mysubfigure{0.45}{0.95}{figures/tomo/weft_tomo_0.png}{CT image $x_{CT}$}{sfig:tomo_weft_tomo_0}
    \caption{Initial XZ slices on the CT and label volumes, the warp yarns are seen longitudinally, reconstruction artifacts are visible in the CT image}%
    \label{fig:unet_tomo_weft_artifacts}
\end{figure}

\begin{figure}
    \centering
    \mysubfigure{0.45}{0.95}{figures/fe/warp_label_070.png}{Label image}{sfig:fe_warp_label_070}
    \mysubfigure{0.45}{0.95}{figures/fe/warp_trinary_070.png}{Trinary conversion}{sfig:fe_warp_trinary_070}
    \par\bigskip

    \mysubfigure{0.45}{0.95}{figures/fe/warp_pseudo_070.png}{Pseudo-CT}{sfig:fe_warp_pseudo_070}
    \mysubfigure{0.45}{0.95}{figures/fe/warp_points_070.png}{Selected points in cross sections}{sfig:fe_warp_points_070}
    \caption{Mid-YZ slices for the 6th simulation step, the weft yarns are seen longitudinally}%
    \label{fig:unet_fe_warp_1}
\end{figure}

\begin{figure}
    \centering
    \mysubfigure{0.45}{0.95}{figures/fe/weft_label_070.png}{Label image}{sfig:fe_weft_label_070}
    \mysubfigure{0.45}{0.95}{figures/fe/weft_trinary_070.png}{Trinary conversion}{sfig:fe_weft_trinary_070}
    \par\bigskip

    \mysubfigure{0.45}{0.95}{figures/fe/weft_pseudo_070.png}{Pseudo-CT}{sfig:fe_weft_pseudo_070}
    \mysubfigure{0.45}{0.95}{figures/fe/weft_points_070.png}{Selected points in cross sections}{sfig:fe_weft_points_070}
    \caption{Mid-XZ slices for the 6th simulation step, the warp yarns are seen longitudinally}%
    \label{fig:unet_fe_weft_1}
\end{figure}

\begin{figure}
    \centering
    \mysubfigure{0.45}{0.95}{figures/fe/warp_label_100.png}{Label image}{sfig:fe_warp_label_100}
    \mysubfigure{0.45}{0.95}{figures/fe/warp_trinary_100.png}{Trinary conversion}{sfig:fe_warp_trinary_100}
    \par\bigskip

    \mysubfigure{0.45}{0.95}{figures/fe/warp_pseudo_100.png}{Pseudo-CT}{sfig:fe_warp_pseudo_100}
    \mysubfigure{0.45}{0.95}{figures/fe/warp_points_100.png}{Selected points in cross sections}{sfig:fe_warp_points_100}
    \caption{Mid-YZ slices for the last simulation step, the weft yarns are seen longitudinally}%
    \label{fig:unet_fe_warp_2}
\end{figure}

%% file: 4_detectron/main.tex
\subsection{Neural network architecture}
\input{1_architecture}

\subsection{Training of the network}
\input{2_training}

\subsection{Implementation details}
\input{3_implementation}

\subsection{Results}
\input{4_results}

%% file: 4_detectron/1_architecture.tex
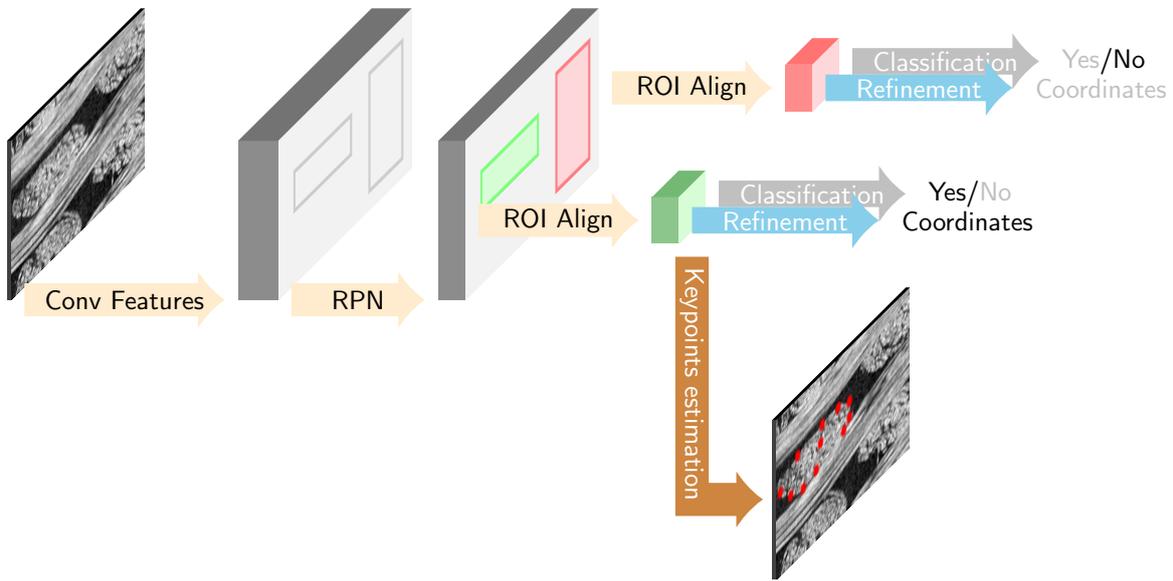
\begin{figure}
    \newcommand{\boxfigure}[7]{%
        \node[#1, yslant=1, inner sep=0, outer sep=0] (img#2) {\includegraphics[width=#3, height=#4]{figures/mask_rcnn/step_#2.png}};%
        \node[minimum width=#5, minimum height=#4, fill=#6, inner sep=0, outer sep=0, anchor=south east] at (img#2.south west) (side#2) {};%
        \node[minimum width=#5, minimum height=#3, fill=#7, inner sep=0, outer sep=0, anchor=south east, xslant=1] at (img#2.north west) (top#2) {};%
    }
    \centering
    \begin{tikzpicture}[auto, node distance=8em,
        arrow/.style={->, -{Latex[length=1em,width=2.25em]}, line width=1.25em, shorten >=0.5em, shorten <=0.5em},
        arrow small/.style={line width=1em, shorten >=0.5em, shorten <=0.5em},
        ]
        \boxfigure{}{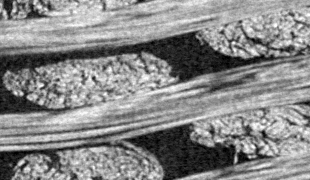}{5em}{6em}{0.15em}{Black!75}{Black};
        \boxfigure{right=of img1.south, anchor=south, xshift=+2em}{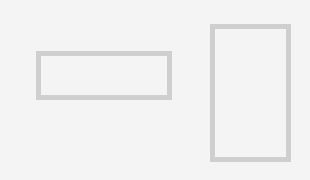}{5em}{6em}{1.5em}{Black!45}{Black!55};
        \boxfigure{right=of img2.south, anchor=south, xshift=-1em}{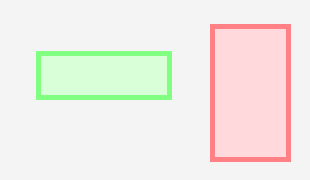}{5em}{6em}{1em}{Black!45}{Black!55};
        \boxfigure{right=of img3.east, anchor=west}{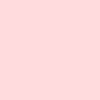}{1.0em}{1.75em}{1em}{Red!45}{Red!55};
        \boxfigure{right=of img3.west, anchor=west}{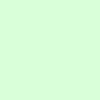}{1.0em}{1.75em}{1em}{Green!45}{Green!55};
        \boxfigure{below right=of img5.south, anchor=north, yshift=+1em}{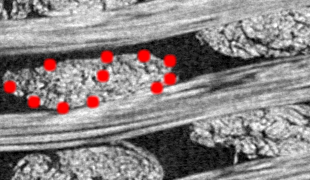}{5em}{6em}{0.15em}{Black!75}{Black};
        \draw[arrow, BlanchedAlmond] (img1.south west) -- node[black, anchor=center] {\textsf{Conv Features}} (side2.south west);
        \draw[arrow, BlanchedAlmond] (img2.south west) -- node[black, anchor=center] {\textsf{RPN}} (side3.south west);
        \draw[arrow, BlanchedAlmond] (img3.east) -- node[black, anchor=center] {\textsf{ROI Align}} (side4);
        \draw[arrow, BlanchedAlmond] (img3.west) -- node[black, anchor=center] {\textsf{ROI Align}} (side5);
        \draw[arrow, Peru, shorten <=1em] (img5.south) |- node[white, rotate=-90, anchor=east, xshift=+1em] {\textsf{Keypoints estimation}} (img6.west);
        \node[right=of img4.east] (img4a) {\textsf{\textcolor{Black!25}{Yes}/No}};
        \node[right=of img4.west] (img4b) {\textsf{\textcolor{Black!25}{Coordinates}}};
        \node[right=of img5.east] (img5a) {\textsf{Yes/\textcolor{Black!25}{No}}};
        \node[right=of img5.west] (img5b) {\textsf{Coordinates}};
        \draw[arrow, arrow small, Silver] (img4.east) -- node[white, anchor=center] {\textsf{Classification}} (img4a);
        \draw[arrow, arrow small, SkyBlue] (img4.west) -- node[white, anchor=center] {\textsf{Refinement}} (img4b);
        \draw[arrow, arrow small, Silver] (img5.east) -- node[white, anchor=center] {\textsf{Classification}} (img5a);
        \draw[arrow, arrow small, SkyBlue] (img5.west) -- node[white, anchor=center] {\textsf{Refinement}} (img5b);

    \end{tikzpicture}
    \caption{Diagram of the Mask R-CNN architecture~\cite{mask_rcnn}}
    \label{fig:mask_rcnn}
\end{figure}

\revision{We propose to solve the problem of yarn segmentation as one of keypoint estimation.
As such, we employ the state-of-the-art Mask-RCNN~\cite{mask_rcnn} architecture that specializes in instance segmentation.}
This implies that a first task is to detect all yarns present in an image by defining their respective bonding boxes.
Then, for each of these detected objects, the second task is to predict a fixed number of keypoints that describe each individual cross section.
Currently, the most-performing approach for the first task is a two-step one: first several bounding box proposals are generated, and then each of these are classified as either containing a relevant object or not.
A schematic representation of these steps is shown in~\cref{fig:mask_rcnn}.

\subsubsection{Convolutional Features}

The first step consists in computing some features on the image.
This is done using a convolutional neural network (CNN), more precisely a fully convolutional network (FCN).
This calculation is very fast since it computes features for several regions of the image while sharing computation between neighboring regions.
In addition, by using a pre-trained FCN (\ie the backbone), one can get a very powerful feature extractor.
We denote this network as $f_{backbone}$ and the features it generates as $h = f_{backbone}(u)$ for any input image $u$.
Here we use a ResNet~\cite{He_2016_ResNet} pre-trained on ImageNet~\cite{Russakovsky_2015_ImageNet} as backbone.

\subsubsection{Region Proposal Network}

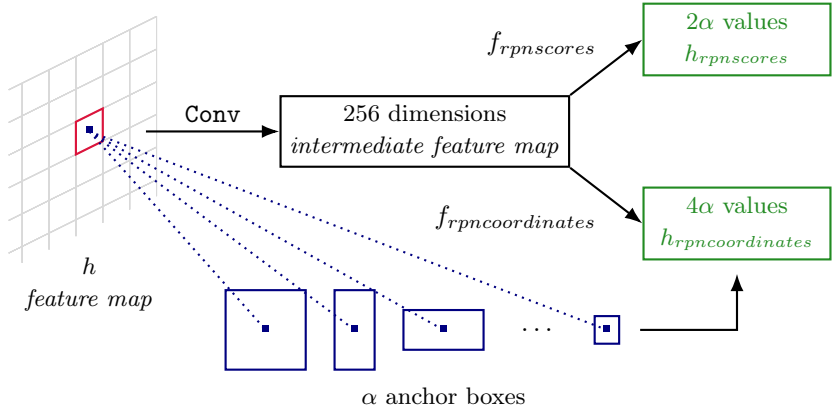
\begin{figure}
    \tdplotsetmaincoords{60}{45}
    \tdplotsetrotatedcoords{0}{0}{0}
    \newcommand*\smallsquare{\mathbin{\vcenter{\hbox{\rule{0.25em}{0.25em}}}}}
    \centering
    \begin{tikzpicture}[auto, thick, node distance = 1em,
        arrow/.style={-Latex},
        tdplot_rotated_coords,
        every text node part/.style={align=center}]
        \foreach \y in {-1.25,-0.75,...,1.25}
        \foreach \z in {-1.25,-0.75,...,1.25}
        {
            \draw[very thin, Black!15] (0, \y, -1.5) -- (0, \y, 1.25);
            \draw[very thin, Black!15] (0, -1.5, \z) -- (0, 1.25, \z);
        }
        \draw[Crimson] (0,-0.25,-0.25) -- (0,0.25,-0.25) -- (0,0.25,0.25) -- (0,-0.25,0.25) -- cycle;
        \node[] at (0,0,0) (n_conv_map) {};
        \node[right=of n_conv_map] (n_conv_map_aux) {};
        \node [draw, right=of n_conv_map_aux, xshift=4em] (n_intermediate) {\small{256 dimensions} \\ \small{\textit{intermediate feature map}}};
        \node [draw, ForestGreen, above right=of n_intermediate, xshift=2em, minimum width=7em] (n_scores) {\small{$2 \alpha$ values} \\ \small{$h_{rpnscores}$}};
        \node [draw, ForestGreen, below right=of n_intermediate, xshift=2em, minimum width=7em] (n_coords) {\small{$4 \alpha$ values} \\ \small{$h_{rpncoordinates}$}};
        \node[below=of n_coords, yshift=1.25em] (n_coords_aux) {};
        \node[below =of n_coords_aux, xshift=-4em, yshift=-0.75em] (box_5_aux) {};
        \node[draw, NavyBlue, left=of box_5_aux, xshift=1em] (box_5) {$\smallsquare$};
        \node[left=of box_5, minimum size=1.5em] (box_4) {\ldots};
        \node[draw, NavyBlue, left=of box_4, minimum width=3.0em, minimum height=1.5em] (box_3) {$\smallsquare$};
        \node[draw, NavyBlue, left=of box_3, minimum width=1.5em, minimum height=3.0em] (box_2) {$\smallsquare$};
        \node[draw, NavyBlue, left=of box_2, minimum size=3.0em] (box_1) {$\smallsquare$};
        \node[below=of n_intermediate] (n_boxes) {};
        \node[below=of box_3] () {\small{$\alpha$ anchor boxes}};
        \draw[arrow] (n_conv_map_aux) -- node [above] {\texttt{Conv}} (n_intermediate);
        \draw[arrow] (n_intermediate.north east) -- node [above left] {$f_{rpnscores}$} (n_scores.west);
        \draw[arrow] (n_intermediate.south east) -- node [below left] {$f_{rpncoordinates}$} (n_coords.west);
        \draw[dotted, NavyBlue] (n_conv_map.center) -- (box_1.center);
        \draw[dotted, NavyBlue] (n_conv_map.center) -- (box_2.center);
        \draw[dotted, NavyBlue] (n_conv_map.center) -- (box_3.center);
        \draw[dotted, NavyBlue] (n_conv_map.center) -- (box_5.center);
        \draw[arrow] (box_5_aux) -| (n_coords_aux);
        \node[NavyBlue] at (n_conv_map) {$\smallsquare$};
        \node[below=of n_conv_map, yshift=-3em] () {\small{$h$} \\ \small{\textit{feature map}}};
    \end{tikzpicture}
    \caption{\revision{Diagram of the Region Proposal Network~\cite{faster_rcnn}}}
    \label{fig:rpn}
\end{figure}

The Region Proposal Network (RPN), as its name suggest, is a CNN whose task is to propose regions.
To do this, the network evaluates a fixed set of possible regions for every spatial unit on the convolutional feature map $h$ provided by the backbone network $f_{backbone}$.
These fixed regions are known as \emph{anchors} and are just bounding boxes with a predefined scale and aspect ratio, as shown in~\cref{fig:rpn}.
Then, the RPN will apply a \texttt{Conv} layer that will output an intermediate feature map (with 256 dimensions at every spatial unit in our case).
This feature map is then fed through two independent branches $f_{rpnscores}$ and $f_{rpncoords}$ so as to obtain the scores $h_{scores} = f_{rpnscores}(h)$ and coordinates $h_{coords} = f_{rpncoords}(h)$.
The score in $h_{scores}$ represents the probability that one anchor contains an object or background (\ie 2 parameters since encoded as a one-hot vector).
The coordinates in $h_{coords}$ represent the transformation that needs to be applied to a given anchor so that it better contains the object.
This transformation typically depicts 2D translation and 2D scaling (\ie 4 parameters).
\revision{As every spatial unit in the feature map $h$ is given the same treatment, for a given set of $\alpha$ anchors, $2\alpha$ scores and $4\alpha$ coordinates will be produced per unit.}

\subsubsection{ROI Align}

\begin{figure}
    \centering
    \begin{subfigure}[b]{0.26\linewidth}
        \centering
        \begin{tikzpicture}
            \node [anchor=south west, inner sep=0] (image) at (0,0)%
                {\includegraphics[width=0.9\linewidth]{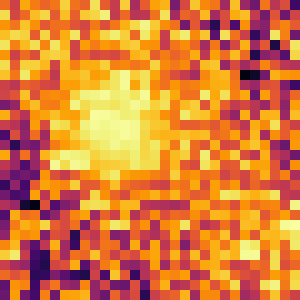}};
            \begin{scope}[x={(image.south east)},y={(image.north west)}, arrow/.style={-Latex}]
                \draw [line width=0.2em, Emerald] (0.18, 0.35) rectangle +(0.70, 0.46);
                \draw [line width=0.2em, Aquamarine] (0.03, 0.03) rectangle +(0.35, 0.46);
            \end{scope}
        \end{tikzpicture}
        \caption{}
        \label{sfig:feature_map}
    \end{subfigure}
    \begin{subfigure}[b]{0.09\linewidth}
        \newcommand{\nodefigure}[1]{\includegraphics[width=0.85\linewidth]{figures/#1.png}}
        \centering
        \begin{subfigure}[b]{\linewidth}
            \centering
            \begin{tikzpicture}
                \node [draw, inner sep=0, line width=0.4em, Emerald] {\nodefigure{roi_align/roi_1_avg}};
            \end{tikzpicture}
        \caption{}
        \label{sfig:roi_align_feature_roi_1_avg}
        \end{subfigure}
        \par\bigskip

        \begin{subfigure}[b]{\linewidth}
            \centering
            \begin{tikzpicture}
                \node [draw, inner sep=0, line width=0.4em, Aquamarine] {\nodefigure{roi_align/roi_2_max}};
            \end{tikzpicture}
        \caption{}
        \label{sfig:roi_align_feature_roi_2_max}
        \end{subfigure}
    \end{subfigure}
    \caption{\revision{\subref{sfig:feature_map} Sample feature map and
    sample extracted feature maps using ROI align.
    The resulting maps of equal size are obtained using \subref{sfig:roi_align_feature_roi_1_avg} average- and
    \subref{sfig:roi_align_feature_roi_2_max} max-pooling respectively}}%
    \label{sfig:roi_align_feature_map}%
\end{figure}

For all regions classified as containing an object by $h_{scores}$ in the previous stage, the information provided in $f_{rpncoords}$ along with the anchor type allow determining the ROI inside the feature map.
However, all selected ROI will not have the same size (\ie different anchors) nor be aligned with the spatial grid (\ie sub-pixel precision).
Then, a technique called ROI align~\cite{mask_rcnn} is used to warp them into a fixed size (\eg $14 \times 14$) without loosing information about their spatial positioning.
This approach essentially consists in performing bilinear interpolation on the impacted region so as obtain an intermediate image that is a multiple of the desired size (\eg $28 \times 28$).
Then a pooling operation (\eg $2 \times 2$~max-pooling) provides the final image with the desired size.
We denote this operation as $f_{align}$ and its output $h_{align} = f_{align}(h_{ROI})$ for any one region $h_{ROI}$ in $h$ identified as containing an object by $h_{scores}$.
An example of this process is shown in~\cref{sfig:roi_align_feature_map}.

\subsubsection{\revision{Bounding Box Classification and Refinement}}

These newly obtained fixed-sized ROI are now employed for object classification and bounding box refinement.
The first part consists in assigning one of multiple classes to the ROI.
Given that in this case we only concern with one class of object, each ROI will be classified as containing either a \enquote{yarn object} or a \enquote{background object} (\eg longitudinally seen yarns).
This operation differs to the one in RPN where the classification was used to decide if an object was present or not.
This classification task is performed with a simple fully connected layer (with 2 output units) $f_{cls}$, and its output is $h_{cls} = f_{cls}(h_{align})$.
The second part employs another fully connected layer (with 4 output units) $f_{box}$ to provide $h_{box} = f_{box}(h_{align})$ which contains the coordinates refinement values.

\subsubsection{Keypoints estimation}

One of the major improvement that Mask-RCNN~\cite{mask_rcnn} brought was to introduce a parallel branch to the Box Classification and Refinement stage that was dedicated to the instance segmentation task.
This parallel branch aims to predict the position of $N_{kp} = 11$ keypoints for each instance.
It consists in some \texttt{Conv} layers followed by an \texttt{Upsample} operation to obtain fixed-sized feature maps $S$ for each keypoint (\eg $11 \times 56 \times 56$).
Each feature map will represent the probability distribution of the localization of the corresponding keypoint.
Then, the 2D coordinates of each keypoint are obtained by taking the position with maximum score for each of the feature maps.

%% file: 4_detectron/2_training.tex
First, it may be useful to recall the definitions of
the Huber loss $\mathcal{H}_{\delta}$, commonly used in robust regression thanks to its \enquote{soft} $L_{1}$ norm being less sensitive to outliers,
the binary cross entropy loss $H_{2}$, commonly used in binary classification,
and the cross entropy loss $H$ for discrete probability distributions.
These are:
\begin{align}
    \mathcal{H}_{\delta}(w) &= \begin{cases}
        \frac{1}{2} w^{2}  & \text{if } \vert w \vert \leq \delta \\
        \vert w \vert \delta - \frac{1}{2} \delta^{2} & \text{otherwise}
    \end{cases} \\
    H_{2}(p, q) &= - q\log(p) - (1 - q)\log(1 - p) \\
    H(p, q) &= -\sum_{\forall x} q(x) \log p(x)
\end{align}
Also, it should be noted that, the coordinates $q_{coords}$ defining the bounding boxes for each yarn are computed as a pre-processing step.
These are easily obtained by extracting the minimum and maximum values of the keypoints coordinates of every yarn.

Training of the networks $f_{backbone}$, $f_{rpncoords}$ and $f_{rpnscores}$ is achieved by optimizing the weighted sum of a classification loss $\mathcal{L}_\text{class}$ and a regression loss $\mathcal{L}_\text{box}$.

The \emph{classification loss} is defined as
\begin{equation}
    \mathcal{L}_\text{class} (p_{i}, q_{i}) = H_{2}(p_{i}, q_{i})
\end{equation}
where $i \in [1, \alpha]$ is the index of the considered anchors,
$p_{i}$ is the score given by the network for that particular anchor (obtained from $h_{scores}$),
and the ground truth (GT) label $q_{i}$ is the boolean result (\ie 1 or 0) of the operation
${\text{IoU}(\text{anchor}, q_{coords}) > \tau}$,
with the intersection over union (IoU, also known as Jaccard index) computed between the corresponding anchor and the respective bounding box,
and a given threshold $\tau$.
The optimization of this loss will encourage the RPN to correctly distinguish between anchors that contain objects and those that do not.

Then, the \emph{regression loss} is defined as
\begin{equation}
    \mathcal{L}_\text{box} = \mathcal{H}_{1}(h_{coords} - q_{coords})
\end{equation}
where $h_{coords}$ represents the coordinates (4 parameters) predicted by the network,
and $q_{coords}$ represents the GT coordinates for the bounding boxes.
The optimization of this loss will tune the Box Classification and Refinement stage so that the anchors correctly include the objects (\ie yarns).
This loss is only computed for anchors that have been labeled as positives.

Next, training of the keypoint estimation branch is performed by optimizing the loss function
\begin{equation}
    \mathcal{L}_{kpt}(S, S_{coords}) = H(S, S_{coords})
\end{equation}
where $S$ represents the distributions estimated by the network for each keypoint in the form of feature maps (\eg $11 \times 56 \times 56$),
and $S_{coords}$ represents the GT distribution for the known keypoints.
This latter is computed for each keypoint as a one-hot vector of dimension 3136 reshaped into a $56 \times 56$ map.
As such, the keypoint estimation task becomes a classification problem where the objective is to determine for each keypoint what position it belongs to (out of the 3136 possible locations).

Finally, training is performed on the pseudo-CT images $I_{pCT}$ generated in the previous section and the corresponding keypoints obtained from FE (see~\cref{sfig:fe_warp_points_070,sfig:fe_weft_points_070,sfig:fe_warp_points_100,sfig:fe_weft_points_100}).
All 33624 slices ($(1401 + 1401) \cdot 12$) are used for training of the networks.

%% file: 4_detectron/3_implementation.tex
The open source project Detectron2~\cite{wu2019detectron2} was used to implement this part of our system.
This software system is also powered by the PyTorch~\cite{paswke_2019_pytorch} framework and implements many state-of-the-art object detection algorithms, such as the Mask R-CNN used here.
This implementation works almost out-of-the-box with the default parameters, and only required adapting the learning rate to a smaller value of $2.5 \times 10^{-5}$.
However, a more substantial modification is required to modify the default number of 17 keypoints used for human pose estimation into the $N_{kp} = 11$ keypoints used for describing the yarns (10 for the cross section and 1 for the yarn path).

\begin{figure}
    \centering
    \begin{tikzpicture}
        \begin{axis}[
            scale only axis=true, cycle list/Set1,
            width=0.4\linewidth, height=0.2\linewidth,
            xmin=0, xmax=1, ymin=0.012, ymax=0.67,
            grid=both, xticklabels={,,}, scaled y ticks=false,
            ]
            \addplot+ [ultra thick, mark=none] table [x=step, y=loss_class, col sep=comma] {4_detectron/figures/losses.csv};
            \addplot+ [ultra thick, mark=none] table [x=step, y=loss_box, col sep=comma] {4_detectron/figures/losses.csv};
            \addplot+ [ultra thick, mark=none] table [x=step, y=loss_kpt, col sep=comma] {4_detectron/figures/losses.csv};
            \addlegendentry{$\mathcal{L}_\text{class}$};
            \addlegendentry{$\mathcal{L}_\text{box}$};
            \addlegendentry{$\mathcal{L}_{kpt}$};
        \end{axis}
    \end{tikzpicture}
    \caption{Loss evolution for the different losses considered}
    \label{fig:maskrcnn_loss_evolution}
\end{figure}
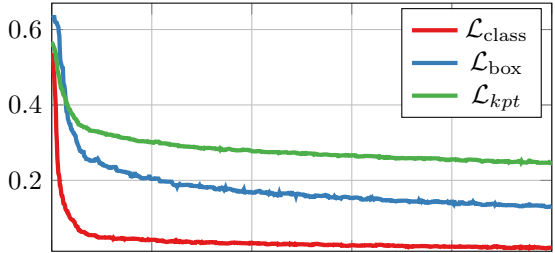

Finally, the model was trained on 1 GPU Nvidia Quadro P4000 (8Gb RAM). 
The loss evolution for the three losses detailed in the previous section is shown in~\cref{fig:maskrcnn_loss_evolution}.
Clearly, $\mathcal{L}_\text{class}$ is the first to attain a lower-valued plateau, followed by $\mathcal{L}_\text{box}$ and $\mathcal{L}_{kpt}$.
These latter two however, continue improving (albeit slowly) at the moment training was stopped.

%% file: 4_detectron/4_results.tex
\begin{figure}
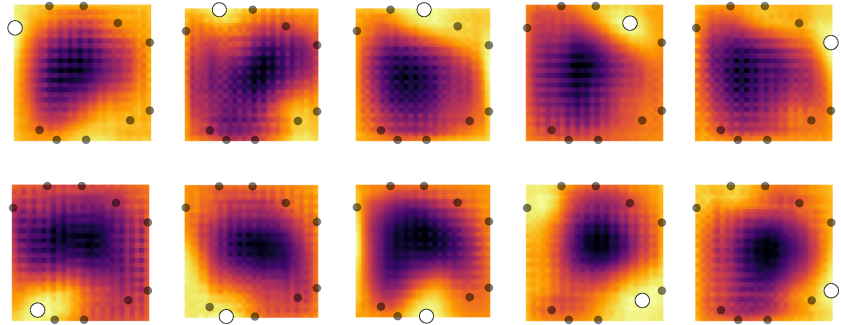

    \newcommand{\mysubfiguresimple}[3]{%
        \begin{subfigure}[b]{#1\linewidth}%
            \centering\includegraphics[width=#2\linewidth]{#3}%
        \end{subfigure}%
    }
    \centering
    \mysubfiguresimple{0.13}{0.90}{figures/heatmaps/3.png}
    \mysubfiguresimple{0.13}{0.90}{figures/heatmaps/4.png}
    \mysubfiguresimple{0.13}{0.90}{figures/heatmaps/5.png}
    \mysubfiguresimple{0.13}{0.90}{figures/heatmaps/6.png}
    \mysubfiguresimple{0.13}{0.90}{figures/heatmaps/7.png}
    \par\bigskip
    
    \mysubfiguresimple{0.13}{0.90}{figures/heatmaps/1.png}
    \mysubfiguresimple{0.13}{0.90}{figures/heatmaps/2.png}
    \mysubfiguresimple{0.13}{0.90}{figures/heatmaps/10.png}
    \mysubfiguresimple{0.13}{0.90}{figures/heatmaps/9.png}
    \mysubfiguresimple{0.13}{0.90}{figures/heatmaps/8.png}
    \caption{Score maps predicted for 10 keypoints of one ROI,
    the white dots indicate the position with maximum score for each feature map,
    and the gray dots do the same but for all other keypoints}
    \label{fig:keypts_maps}
\end{figure}
\begin{figure}
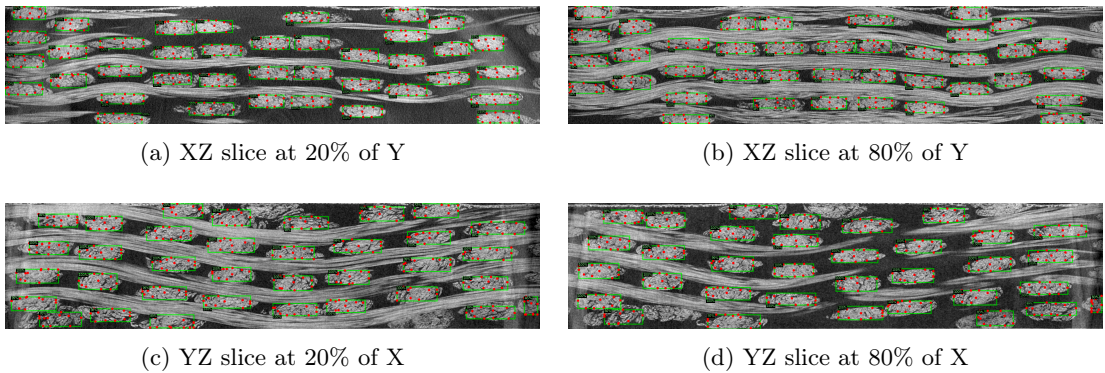

    \centering
    \mysubfigure{0.45}{0.95}{figures/results/warp_out_1.png}{XZ slice at 20\% of Y}{sfig:warp_out_1}%
    \mysubfigure{0.45}{0.95}{figures/results/warp_out_3.png}{XZ slice at 80\% of Y}{sfig:warp_out_3}%
    \par\bigskip

    \mysubfigure{0.45}{0.95}{figures/results/weft_out_1.png}{YZ slice at 20\% of X}{sfig:weft_out_1}%
    \mysubfigure{0.45}{0.95}{figures/results/weft_out_3.png}{YZ slice at 80\% of X}{sfig:weft_out_3}%
    \caption{Visualization of the instance segmentation results for some slices of the acquired CT image,
    bounding box prediction shown in green and
    keypoints estimation as red dots (inside each bounding box)}%
    \label{fig:detection_results}
\end{figure}

After training on the numerical dataset (from FE), the network can be used for predicting on the acquired CT volume $I_{CT}$.
First, an example of the predicted score maps for some keypoints is shown in~\cref{fig:keypts_maps}.
As it can be seen, the keypoints coordinates can be directly extracted by identifying the position at which the maximum score value occurs.
Then, \cref{fig:detection_results} displays the results of both stages, bounding box detection and keypoint estimation.
The system performs remarkably well in the detection stage, as it shows no false positives (false yarn detection) but some few false negatives (missing yarn detection).
An example of this can be seen in~\cref{sfig:weft_out_3}, where a a yarn is not detected on the bottom left part of the image.
This issue will be further explored in the next section.

%% file: 5_meshing/main.tex
\subsection{Completion of missing sections}
\label{subsec:completion_missing_sections}
\input{1_completion}

\subsection{Mesh of yarn path}
\label{subsec:mesh_yarn_path}
\input{2_mesh_yarn_path}

\subsection{Mesh of reinforcement}
\label{subsec:mesh_reinforcement}
\input{3_mesh_reinforcement}

\subsection{Mesh of composite}
\label{subsec:mesh_composite}
\input{4_mesh_composite}

\begin{figure}
    \centering
    \begin{subfigure}[b]{0.45\linewidth}
        \centering
        \begin{tikzpicture}
            \node [anchor=south west, inner sep=0] (image) at (0,0)%
                {\includegraphics[width=0.73\linewidth]{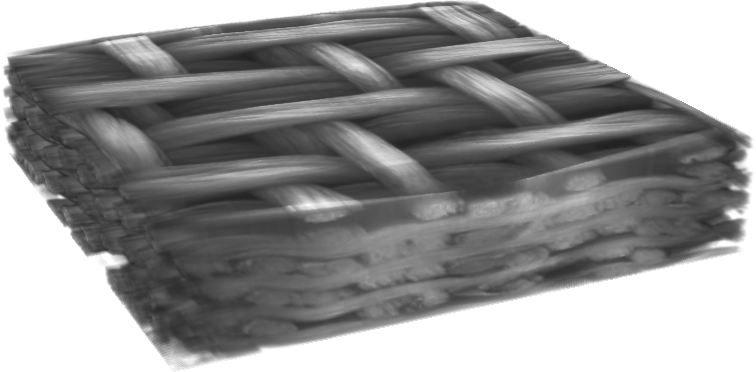}};
            \begin{scope}[x={(image.south east)},y={(image.north west)}, arrow/.style={-Latex}]
                \node at (1.03, 0.64) (px) {$x$};
                \node at (-0.03, 0.89) (py) {$y$};
                \draw [arrow] (0.62, 1.05) -- (px);
                \draw [arrow] (0.62, 1.05) -- (py);
            \end{scope}
        \end{tikzpicture}
        \caption{}
        \label{sfig:original_tomo}
    \end{subfigure}
    \mysubfigure{0.45}{0.75}{figures/mesh_1.png}{}{sfig:mesh_evolution_1}%
    \par\bigskip

    \mysubfigure{0.45}{0.75}{figures/mesh_2.png}{}{sfig:mesh_evolution_2}%
    \mysubfigure{0.45}{0.75}{figures/mesh_3.png}{}{sfig:mesh_evolution_3}%
    \caption{
    \subref{sfig:original_tomo} Acquired CT volume $I_{CT}$ of the studied unit cell,
    and resulting FE meshes for
    \subref{sfig:mesh_evolution_1} the yarn path,
    \subref{sfig:mesh_evolution_2} the reinforcement (surface or volume), and
    \subref{sfig:mesh_evolution_3} the composite (volume)}
    \label{fig:mesh_evolution}
\end{figure}

%% file: 5_meshing/1_completion.tex
Since no consistency is enforced between slices, in some cases a yarn cross section may be detected in a given slice but not detected in the subsequent, and vice versa.
Hence a simple procedure is proposed for \enquote{filling} the missing sections.

Given that the number of warp and weft yarns is known (39 warp and 36 weft), a simple look-up on the number of detected instances per slice will yield the \enquote{incomplete} slices (those with less yarn cross sections than expected) as well as the complete ones.
Indeed, from the 3512 analyzed slices (1814 XZ slices and 1698 YZ slices), 37 (1\%) present one more instance than excepted and only 25\% of the total contain the exact number of instances.
As~\cref{table:segmentation_instances} shows, most slices (62\%) are only missing one or two instances, with the warp direction being more precise than the weft one.
Thus, 95\% of all instances (all cross sections for all yarns for all slices) are correctly detected by the neural network.

It also is important to note that the missing yarn cross sections in those incomplete slices are in their majority (84\%) from the upper-most and lower-most layers of the textile (\ie the boundaries).
Similarly, most of the cases in which a yarn cross section is missing in a large number of consecutive slices only happen at the lateral extremes of the volume, region in which the tomographic artifact is most prominent (see~\cref{sfig:tomo_warp_tomo_0}).
\revision{It should be noted that the Mask R-CNN was not trained on images presenting tomographic artifacts because the U-Net does not generate them.
Thus, the network is prone to under-perform in this regions.}

After simple examination of the predicted keypoints it became clear that the center keypoint often was not placed in the \enquote{center} of the yarn.
Let us recall that no particular attention was brought to this center keypoint, rather all 11 were treated equally.
\revision{Hence, the center keypoint will be recomputed as the centroid from the correctly predicted 10 contour points.}
These new center keypoints are used hereinafter.

Then, all pairs of subsequent slices that are composed of one complete and one incomplete slice are identified by computing the pair-wise distance matrix between the computed center keypoints present in both.
This matrix helps pairing each yarn in the incomplete slice to the correct yarn in the complete slice by observing their distances.
Next, all 11 keypoints for the unpaired yarns in the complete slice are simply copied onto the incomplete one.
This procedure is repeated for all incomplete slices until none is left.
It was observed that this naive \enquote{copy} mechanism was effective in sight of the good consistency of the network and the almost non existent complicated cases (\eg a long portion of a missing yarn).

\begin{table}
    \centering
    \caption{
    Segmentation assessment using
    \subref{stable:missing_instances} the number of missing instances per slice that require completion, and
    \subref{stable:detected_instances} the number of detected instances in the whole volume.
    }%
    \label{table:segmentation_instances}
    \begin{subfigure}[t]{0.49\linewidth}
    \centering
    \caption{}
    \label{stable:missing_instances}
    \begin{tabular}{@{}lrrr@{}}
        \toprule
        & Warp & Weft & Both \\
        \midrule
        One           & 42\% & 30\% & 36\% \\
        Two           & 11\% & 39\% & 26\% \\
        Three or more &  3\% & 21\% & 12\% \\
        \bottomrule
    \end{tabular}
    \label{table:missing_instances}

    \end{subfigure}
    \begin{subfigure}[t]{0.49\linewidth}
    \centering
    \caption{}
    \label{stable:detected_instances}
    \begin{tabular}{@{}lrrr@{}}
        \toprule
        & Warp & Weft & Both \\
        \midrule
        All         & 66222 &  65304 &  131526 \\
        Detected    & 64269 &  60515 &  124784 \\
        \textbf{\% Detected} & \textbf{97\%} & \textbf{93\%} & \textbf{95\%} \\
        \bottomrule
    \end{tabular}
    \label{table:detected_instances}

    \end{subfigure}
    
\end{table}

%% file: 5_meshing/2_mesh_yarn_path.tex
As it can be seen in~\cref{sfig:yarn_evolution_1}, the identified yarn paths are very ragged as a consequence of per-slice (\ie independent) prediction.
A simple median filtering is applied along each yarn path so as to smoothen them out.
Next, in order to alleviate the meshes to be generated, a sub-sampling procedure is applied.
Then, by retaining only 1 cross section every 25 slices, 66 (out of 1698) for warp yarns and 71 (out of 1814) for weft yarns are kept.
The corresponding meshes (using L1 elements) can be seen in~\cref{fig:yarn_evolution,sfig:mesh_evolution_1}.

\begin{figure}
    \centering
    \mysubfigure{0.45}{0.75}{figures/yarn_1.png}{}{sfig:yarn_evolution_1}%
    \mysubfigure{0.45}{0.75}{figures/yarn_3.png}{}{sfig:yarn_evolution_3}%
    \caption{Identified yarn paths
    \subref{sfig:yarn_evolution_1} after completion of missing cross sections,
    \subref{sfig:yarn_evolution_3} after median filtering and sub-sampling,
    the warp and weft colored in black and gold respectively}
    \label{fig:yarn_evolution}
\end{figure}

%% file: 5_meshing/3_mesh_reinforcement.tex
The generation of a conformal FE mesh for the yarns is a trivial process.
Since the instances provided by the network contain the same number of keypoints describing they yarn cross sections, these can simply be connected.
\revision{First, the (local) polar coordinates of the proposed keypoints are computed for obtaning a consistent order.
Then, all pairs of keypoints in consecutive cross sections are connected (\eg keypoint 1 in section 1 with keypoint 1 in section 2).}
As such, a mesh surface composed of quadrilateral (Q4) elements can be generated for each yarn.
These can be seen in~\cref{sfig:mesh_evolution_2}.

Moreover, these surface elements can be connected to the previous (L1) mesh for the yarn path so as to construct a volume mesh for each yarn using triangular prisms (P6, wedges).

\revision{It should be noted these meshes can be taken directly for simulation purposes, or used as input geometries for a more advanced meshing software.}

%% file: 5_meshing/4_mesh_composite.tex
\begin{figure}
    \centering
    \includegraphics[width=0.30\linewidth]{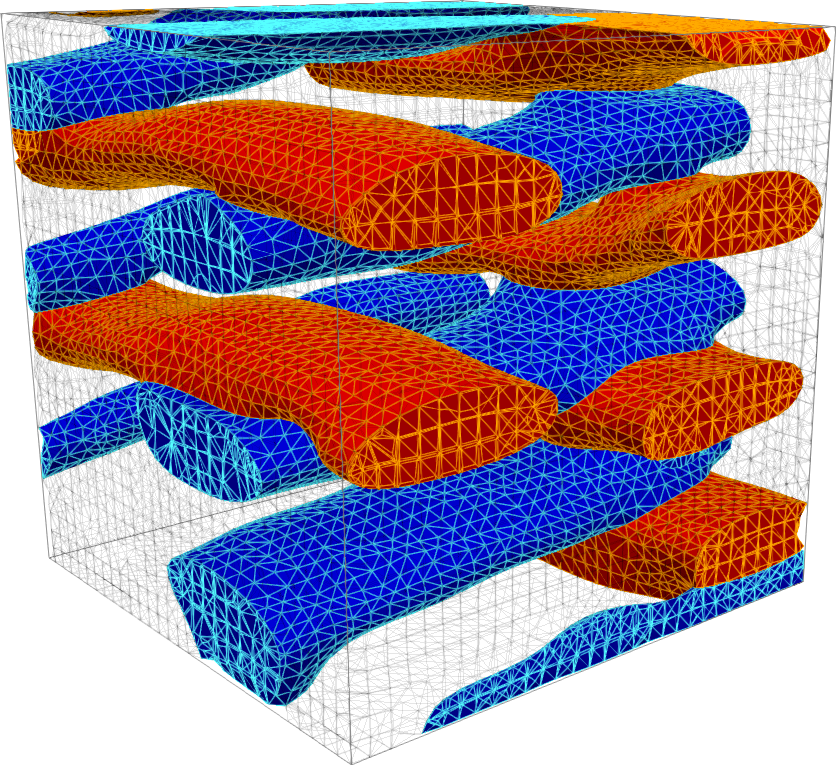}
    \caption{Detailed view on the conformal mesh of the composite: the warp and weft elements are shown in red and blue, while the matrix is shown as a wireframe}
    \label{fig:mesh_3_detail}
\end{figure}

Here, the textile description (yarn path and associated cross sections) is fed to the in-house software REVoxel~\cite{Hello_2014_Numerical} in order to obtain an intermediate voxel-FE mesh.
\revision{This latter is then fed to the in-house software Mirax (developed by Alain Rassineux at UTC) which outputs a fully conformal mesh using tetrahedral (T4) elements.}
The resulting mesh is showed in~\cref{sfig:mesh_evolution_3}, and a detailed view can be seen in~\cref{fig:mesh_3_detail}.

%% file: 6_validation/main.tex
Usually, the validation of segmentation models is performed qualitatively~\cite{Sinchuk_Materials_2020, Ali_CompPartA_2020_Deep}.
However some quantitatively validation can be obtained by extracting relevant features (\eg yarn path, section areas~\cite{Huang_CompPartA_2019}), using similarity metrics~\cite{Benezech_CS_2019} (\ie comparing binary mask, local directions and yarn inter-penetration), or by assessing the mechanical or thermal properties of the textile~\cite{Naouar_CS_2015, Ewert2020}.

In our case, the simple textile model acquired using the semi-automatic procedure described in~\cref{subsec:simple_textile_model} can be used as ground truth for validating the performance of the algorithm.
However, it should be noted that only the yarn paths can be considered as \enquote{true} since the cross sections are simple ellipses.

Hence, the Haussdorf metric will be used for the quantitative evaluation of the yarn paths, while the intra-yarn fiber volume fraction will be used for validation of the cross sections.
Note that the assessment of the instance segmentation is performed on the final meshed model.

\subsection{Path assessment}
\label{subsec:path_assessment}
\subimport*{1_path/}{main}

\subsection{Section assessment}
\label{subsec:section_assessment}
\subimport*{2_textile/}{main}

%% file: 6_validation/1_path/main.tex
In order to provide a quantitative comparison, we propose to use the one-sided Hausdorff distance, a commonly used metric in 3D segmentation tasks defined as:
\begin{equation}
    \delta_{H}(A, B) = \max_{a \in A} \min_{b \in B} \lVert a - b \rVert
\end{equation}
with the sets $A$ and $B$ and $\delta_{H}(A, B)$ measuring the distance from $A$ to $B$. 

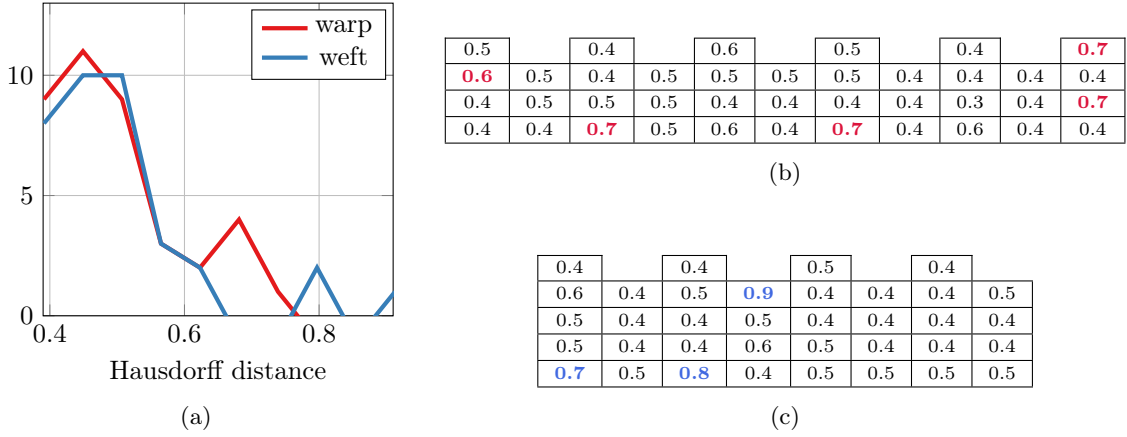
\begin{figure}
    \centering
    \begin{subfigure}[b]{0.35\linewidth}
        \centering
        \begin{tikzpicture}
            \begin{axis}[
                scale only axis=true, grid=both, cycle list/Set1,
                width=0.8\linewidth, height=0.72\linewidth,
                xmin=0.39, xmax=0.91, ymin=0, ymax=13,
                xlabel=Hausdorff distance,
                ]
                \addplot+ [ultra thick, mark=none] table [x index=0, y index=2, col sep=comma] {6_validation/1_path/figures/histogram.csv};
                \addplot+ [ultra thick, mark=none] table [x index=0, y index=3, col sep=comma] {6_validation/1_path/figures/histogram.csv};
                \addlegendentry{warp};
                \addlegendentry{weft};
            \end{axis}
        \end{tikzpicture}
        \caption{}
        \label{sfig:hausdorff_distances_histogram}
    \end{subfigure}
    \newcommand{\ca}[1]{\textcolor{Crimson}{\textbf{#1}}}
    \newcommand{\ce}[1]{\textcolor{RoyalBlue}{\textbf{#1}}}
    \begin{subfigure}[b]{0.55\linewidth}
        \centering
        \begin{subfigure}[b]{\linewidth}
            \centering
            \scriptsize
            \begin{tabular}{*{11}{|c}|}
                \cline{1-1} \cline{3-3} \cline{5-5} \cline{7-7} \cline{9-9} \cline{11-11}
                0.5 &  & 0.4 &  & 0.6 &  & 0.5 &  & 0.4 &  & \ca{0.7} \\ \hline
                \ca{0.6} & 0.5 & 0.4 & 0.5 & 0.5 & 0.5 & 0.5 & 0.4 & 0.4 & 0.4 & 0.4 \\ \hline
                0.4 & 0.5 & 0.5 & 0.5 & 0.4 & 0.4 & 0.4 & 0.4 & 0.3 & 0.4 & \ca{0.7} \\ \hline
                0.4 & 0.4 & \ca{0.7} & 0.5 & 0.6 & 0.4 & \ca{0.7} & 0.4 & 0.6 & 0.4 & 0.4 \\ \hline
            \end{tabular}
            \normalsize
            \caption{}
            \label{sfig:hausdorff_distances_warp}
        \end{subfigure}
        \vspace*{\baselineskip}

        \begin{subfigure}[b]{\linewidth}
            \centering
            \scriptsize
            \begin{tabular}{*{8}{|c}}
                \cline{1-1} \cline{3-3} \cline{5-5} \cline{7-7}
                0.4 &  & 0.4 &  & 0.5 &  & 0.4 &  \\ \hline
                0.6 & 0.4 & 0.5 & \ce{0.9} & 0.4 & 0.4 & 0.4 & \multicolumn{1}{c|}{0.5} \\ \hline
                0.5 & 0.4 & 0.4 & 0.5 & 0.4 & 0.4 & 0.4 & \multicolumn{1}{c|}{0.4} \\ \hline
                0.5 & 0.4 & 0.4 & 0.6 & 0.5 & 0.4 & 0.4 & \multicolumn{1}{c|}{0.4} \\ \hline
                \ce{0.7} & 0.5 & \ce{0.8} & 0.4 & 0.5 & 0.5 & 0.5 & \multicolumn{1}{c|}{0.5} \\ \hline
            \end{tabular}
            \normalsize
            \caption{}
            \label{sfig:hausdorff_distances_weft}
        \end{subfigure}
    \end{subfigure}
    \caption{
    \revision{\subref{sfig:hausdorff_distances_histogram} Histogram of Hausdorff distances (in $mm$) by orientation,
    and listing of all distances for each
    \subref{sfig:hausdorff_distances_warp} warp and
    \subref{sfig:hausdorff_distances_weft} weft yarn
    (each cell represents a yarn).
    Yarns with distance greater than 90\% percentile shown in \textbf{bold}}}
    \label{fig:hausdorff_distances}
\end{figure}

\begin{figure}
    \centering
    \begin{subfigure}[b]{0.40\linewidth}
        \centering
        \begin{tikzpicture}
            \node[anchor=south west, inner sep=0] (image) at (0, 0) {
              \centering
              \includegraphics[width=0.6\linewidth]{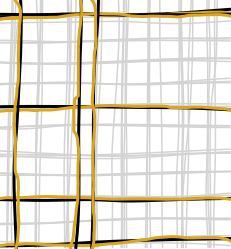}
            };
            \begin{axis}[
                scale only axis, ticks=none,
                width=0.6\linewidth, height=0.65\linewidth,
                xmin=2.5, xmax=35, ymin=2.0, ymax=37,
                xlabel={$x$}, ylabel={$y$}]
            \end{axis}
        \end{tikzpicture}
    \end{subfigure}
    \begin{subfigure}[b]{0.50\linewidth}
        \centering
        \begin{subfigure}[b]{\linewidth}
            \centering
            \begin{tikzpicture}
                \node[anchor=south west, inner sep=0] (image) at (0, 0) {
                  \centering
                  \includegraphics[width=0.70\linewidth]{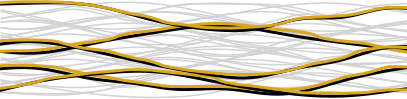}
                };
                \begin{axis}[
                    scale only axis, ticks=none,
                    width=0.70\linewidth, height=0.17\linewidth,
                    xmin=2.5, xmax=35, ymin=0.96106, ymax=8.39594,
                    xlabel={$x$}, ylabel={$z$}]
                \end{axis}
            \end{tikzpicture}
        \end{subfigure}
        \vspace*{\baselineskip}
    
        \begin{subfigure}[b]{\linewidth}
            \centering
            \begin{tikzpicture}
                \node[anchor=south west, inner sep=0] (image) at (0, 0) {
                  \centering
                  \includegraphics[width=0.75\linewidth]{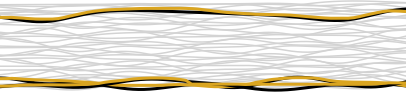}
                };
                \begin{axis}[
                    scale only axis, ticks=none,
                    width=0.75\linewidth, height=0.17\linewidth,
                    xmin=2.0, xmax=37, ymin=0.96106, ymax=8.39594,
                    xlabel={$y$}, ylabel={$z$}]
                \end{axis}
            \end{tikzpicture}
        \end{subfigure}
    \end{subfigure}
    \caption{Comparison between the predicted yarn path in gold and the ground truth in black,
    only showing for yarns with Hausdorff distance greater than 90\% percentile,
    the remaining yarns in light gray}
    \label{fig:hausdorff_bad_yarns}
\end{figure}

In the present case, the measurement is computed all yarns independently.
Thus, for each yarn $i$, the set $A_{i}$ contains the predicted center keypoints (after the smoothing process described previously), and the set $B_{i}$ contains the known keypoints (ground truth).
The obtained distances $\delta_{H}(A_i, B_i)$ for all yarns $i$ can be summarized in the histogram shown in~\cref{sfig:hausdorff_distances_histogram}.
\revision{Moreover, the distance for each yarn is displayed in~\cref{sfig:hausdorff_distances_warp,sfig:hausdorff_distances_weft} following an arrangement that mimics that of the textile (\ie each cell represents a yarn).}
It should become clear that the vast majority of yarns have very small distances which validates the performance of the system.
Some few yarns do have greater distances, but these are rather outliers with very few occurrences.
These are highlighted in~\cref{fig:hausdorff_distances} and properly shown in~\cref{fig:hausdorff_bad_yarns}.
Clearly, the weft orientation is the one that poses the most problems.
It is important to note however, that the three weft yarns responsible for the greatest Hausdorff distances had many consecutive missing slices (as manifested by the straight lines that result from the \enquote{copy} mechanism) and are located on the upper-most and bottom-most layers (as noted in~\cref{subsec:completion_missing_sections}).

%% file: 6_validation/2_textile/main.tex
As said previously the assessment of yarn sections using classical metrics (\eg IoU, F1-score) cannot be done since the simple textile model only considers elliptical cross sections.
Nevertheless, the textile geometrical information, in the form of the intra-yarn fiber volume fraction (FVF), can be used for assessing the quality of the cross section estimation.

More precisely, instead of estimating the average FVF per yarn, we compute the section-wise FVF, or fiber section fraction (FSF), as the ratio between the nominal section and the estimated sections.
The nominal section is computed as $N \cdot \pi \cdot R^{2}$ with $N$ the number of fibers and $R$ their radii, whereas the estimated section is computed as the polygonal area that the keypoints enclose.

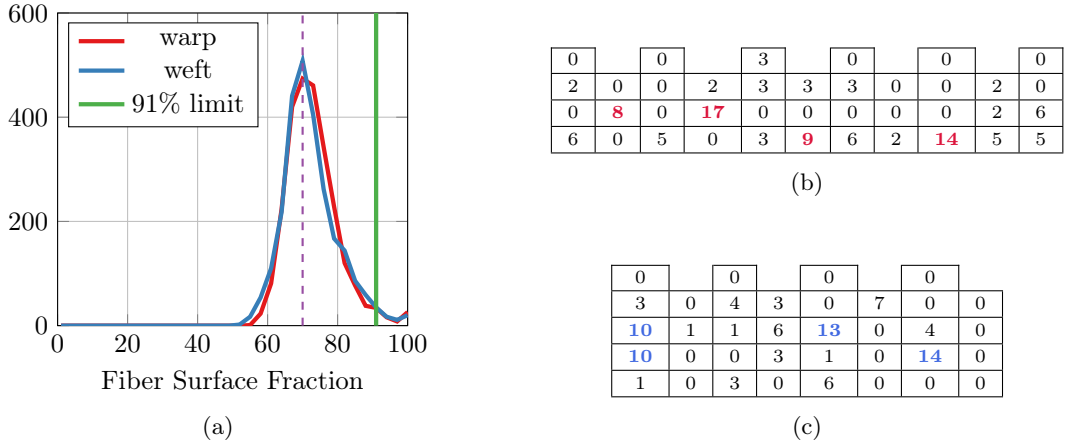
\begin{figure}
    \centering
    \begin{subfigure}[b]{0.35\linewidth}
        \centering
        \begin{tikzpicture}
            \begin{axis}[
                scale only axis=true, grid=both, cycle list/Set1, legend pos=north west,
                width=0.8\linewidth, height=0.72\linewidth,
                xmin=0, xmax=100, ymin=0, ymax=600,
                xlabel=Fiber Surface Fraction
                ]
                \addplot+ [ultra thick, mark=none] table [x index=0, y index=1, col sep=comma] {6_validation/2_textile/figures/histogram.csv};
                \addplot+ [ultra thick, mark=none] table [x index=0, y index=2, col sep=comma] {6_validation/2_textile/figures/histogram.csv};
                \addplot+ [ultra thick, mark=none] coordinates {(91, 0) (91, 600)};
                \addplot+ [thick, dashed, mark=none] coordinates {(70, 0) (70, 600)};
                \addlegendentry{warp};
                \addlegendentry{weft};
                \addlegendentry{91\% limit};
            \end{axis}
        \end{tikzpicture}
        \caption{}
        \label{sfig:fvf_histogram}
    \end{subfigure}
    \newcommand{\ca}[1]{\textcolor{Crimson}{\textbf{#1}}}
    \newcommand{\ce}[1]{\textcolor{RoyalBlue}{\textbf{#1}}}
    \begin{subfigure}[b]{0.55\linewidth}
        \centering
        \begin{subfigure}[b]{\linewidth}
            \centering
            \scriptsize
            \begin{tabular}{*{11}{|c}|}
                \cline{1-1} \cline{3-3} \cline{5-5} \cline{7-7} \cline{9-9} \cline{11-11}
                0 &   & 0 &    & 3 &   & 0 &   & 0  &   & 0 \\ \hline
                2 & 0 & 0 & 2  & 3 & 3 & 3 & 0 & 0  & 2 & 0 \\ \hline
                0 & \ca{8} & 0 & \ca{17} & 0 & 0 & 0 & 0 & 0  & 2 & 6 \\ \hline
                6 & 0 & 5 & 0  & 3 & \ca{9} & 6 & 2 & \ca{14} & 5 & 5 \\ \hline
            \end{tabular}
            \normalsize
            \caption{}
            \label{sfig:fvf_warp}
        \end{subfigure}
        \vspace*{\baselineskip}

        \begin{subfigure}[b]{\linewidth}
            \centering
            \scriptsize
            \begin{tabular}{*{8}{|c}}
                \cline{1-1} \cline{3-3} \cline{5-5} \cline{7-7}
                0 &   & 0 &   & 0 &   & 0 & \\ \hline
                3 & 0 & 4 & 3 & 0 & 7 & 0 & \multicolumn{1}{c|}{0} \\ \hline
                \ce{10} & 1 & 1 & 6 & \ce{13} & 0 & 4 &  \multicolumn{1}{c|}{0} \\ \hline
                \ce{10} & 0 & 0 & 3 & 1 & 0 & \ce{14} & \multicolumn{1}{c|}{0} \\ \hline
                1 & 0 & 3 & 0 & 6 & 0 & 0 & \multicolumn{1}{c|}{0} \\ \hline
            \end{tabular}
            \normalsize
            \caption{}
            \label{sfig:fvf_weft}
        \end{subfigure}
    \end{subfigure}
    \caption{
    \revision{\subref{sfig:fvf_histogram} Histogram of fiber fraction per section (in \%) by orientation with the theoretical limit of 91\%;
    additionally the percentage of sections per yarn that exceed this limit is listed for 
    \subref{sfig:fvf_warp} warp and
    \subref{sfig:fvf_weft} weft yarn
    (each cell represents a yarn)}}
    \label{fig:fvf_summary}
\end{figure}

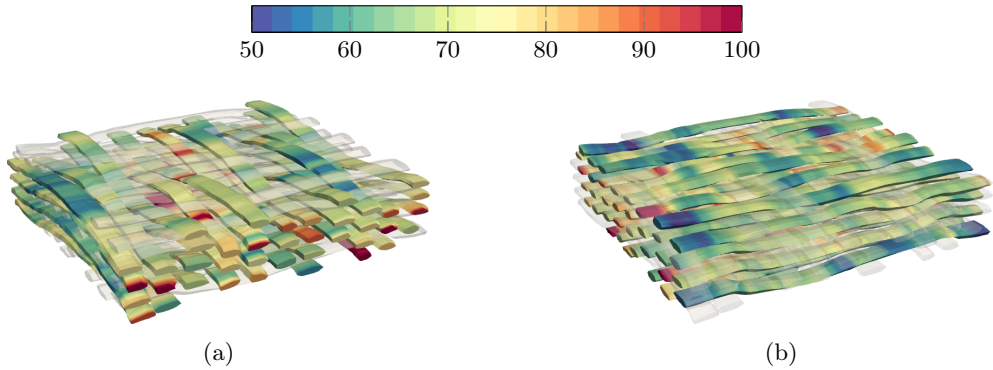
\begin{figure}
    \centering

    \begin{subfigure}[b]{0.49\linewidth}
        \centering
        \begin{tikzpicture}
        \begin{axis} [
            hide axis,
            scale only axis,
            height=0pt,
            width=0pt,
            colorbar horizontal,
            colorbar sampled,
            colorbar style={
                width=0.8\linewidth,
                height=1em,
                tick label style={font=\small},
            },
            colormap/Spectral,
            colormap={reverse Spectral}{
                indices of colormap={
                    \pgfplotscolormaplastindexof{Spectral},...,0 of Spectral}
            },
            point meta min=50,
            point meta max=100,
        ]
        \addplot [draw = none] coordinates {(0,0) (1,1)};
        \end{axis}
        \end{tikzpicture}
    \end{subfigure}
    \par\bigskip
    
    \mysubfigure{0.45}{0.75}{figures/warp_sections.png}{}{sfig:warp_sections}%
    \mysubfigure{0.45}{0.75}{figures/weft_sections.png}{}{sfig:weft_sections}%

    \caption{\revision{Visualization of the FSF spatial distribution for
    \subref{sfig:warp_sections} warp and
    \subref{sfig:weft_sections} weft yarns}}
    \label{fig:fvf_sections}
\end{figure}

\revision{The histogram for all 5130 estimated FSF values ($66 \times 39 + 71 \times 36$) is shown in~\cref{sfig:fvf_histogram} and the spatial distribution is shown in~\cref{fig:fvf_sections}.
It can be seen that for both warp and weft cases, the values are centered around 70\% and range overall from 55 to 90\%. 
These results are in good agreement with similar numerical studies without post-processing (50-95\%)~\cite{Wintiba_CS_2020}, yet experimentally it is difficult to exceed 70–75\%~\cite{Karahan_CompPartA_2010, Green_CS_2014_paper2, Wintiba_CS_2020}.
This global overestimation of the FSF is in part due to a modest but systematic underestimation of the yarn cross section by the network.
However, it is also explained by the method for the calculation of the FSF employed here.
Indeed, a rigorous calculation of the section areas should be done orthogonal to the yarn path (even with low crimp), instead of using the ones aligned with the image axes.
Similarly, a better estimation of sections area using higher order interpolators would prove useful for accurately computing the FSF value.}

\begin{figure}
    \centering
    \includegraphics[width=0.37\linewidth]{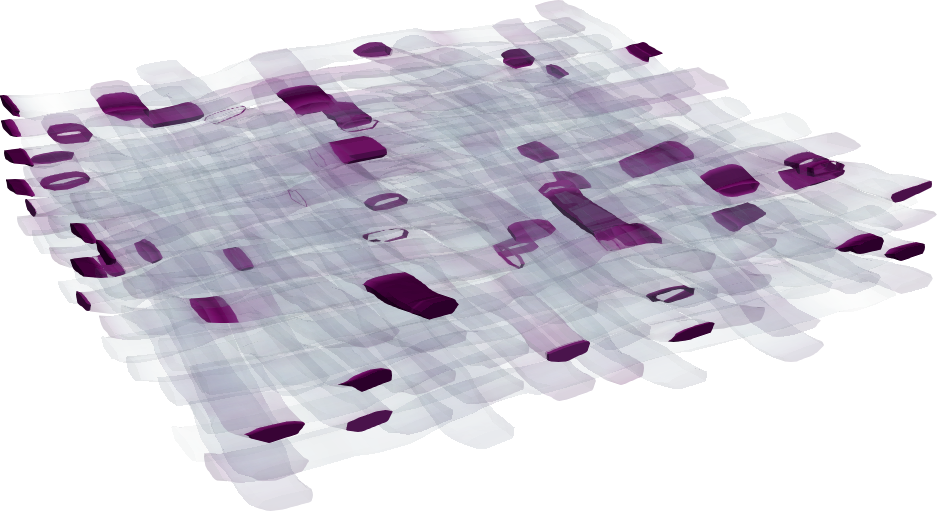}
    \caption{Visualization of sections with FSF higher than the theoretical limit of 91\%}
    \label{fig:fvf_bad_sections}
\end{figure}

\revision{It should be noted that cases in which the estimated section was smaller than the nominal one, the FSF was set to 100\%.
This explains the slight increase observed in the histogram for the highest values close to 100\%.
Moreover, considering that a FSF of 91\% is the theoretical limit for a compact hexagonal arrangement in a 2D plane~\cite{Green_CS_2014_paper2}, sections that surpass this value could be considered as not physically-admissible.}
These constitute 2.6\% of analyzed sections, they are highlighted in~\cref{sfig:fvf_warp,sfig:fvf_weft} and shown in~\cref{fig:fvf_bad_sections}.

%% file: 7_conclusions/main.tex
A new segmentation method based on deep learning has been proposed in this study to predict the yarn envelopes in slices of CT images for 3D woven composites.
It proposes a complete system that includes a first step of generation of synthetic images (U-Net), and a second step of instance segmentation (Mask R-CNN).
These constitute the two main contributions of this study.
Moreover, the fact that a \emph{fully} annotated database is not available, but rather a \emph{weakly} annotated one, is the reason why the first step is crucial.

\revision{We propose to construct a numerical database using the FE mechanical simulations, thus profiting from this substantial knowledge-base and the possibility of performing mechanically sound \emph{data augmentation} by extracting different textile configurations (\eg with variable density).
Moreover, the dataset creation problem is treated as one of \emph{inverse segmentation} from labeled images (issued from FE) into realistic pseudo-CT ones.}
This allows to circumvent the classical problem of requiring a heavily annotated dataset for training a proper instance segmentation model.

The formulation of instance segmentation task used here allows the individual identification of each yarn (\eg by ID) as well as a 10-point parametrization useful for constructing a textile model.
It should be noted that 10 keypoints were found to be sufficient for mechanic elastic homogenization, but the modifications made to the Detectron2 framework (for Mask R-CNN) allow to freely adjust this number of keypoints if required.
Having the system output the yarn keypoints facilitates the use of the readily available meshing strategies (linear, surface, volume) to obtain a description of only the reinforcement or the complete composite.

Then, we assess the results using the Hausdorff distance for the yarn path and the local intra-yarn fiber volume fraction (\ie FSF) for the yarn cross sections.
This two-step validation is dictated by the fact that only a weakly annotated dataset is available.
Additionally, it is a novel strategy that quantifies the results from both, a relative criterion (Hausdorff) as well as an absolute one that represents a material property.
These quantitative validation strategies show an overall good performance while also highlighting areas of improvement.
For example the overestimation of FSF could be introduced as a penalization factor in the loss functions, thus avoiding unrealistically high values (\eg higher than 91\%).
Similarly, the number of keypoints could be adapted according to the yarn or fabric types (\ie higher or lower crimp fabrics).

A more substantial improvement could be to emphasize the center keypoint (\ie differentiate it from the others) so that it truly represents the yarn path.
Alternatively, this keypoint could be completely disregarded and just computed on the fly from the remaining 10 ones.
Another approach would be to use a polar representation (with the origin at the center keypoint) with some constraints on the expected distances between contour keypoints.
This implies a sizable departure from the classical Mask R-CNN and may be explored in future works.

Another area for improvement is the spatial coherence through multiple slices, this problem is explored in~\ref{subsec:consistency}.
Indeed, the 2.5D approach taken here does not impose any connection between consecutive slices (other than them belonging to the same manifold).
The simplest approach is then to adapt existing 2D and temporal frameworks so that the temporal coherence is interpreted as a spatial one for our case.
This would have the great advantage of eliminating the completion step employed here, since the network would not omit nor add yarns sporadically.
A more complex strategy would be to conceive a fully 3D system that would deal with 3D sub-volumes instead of slices.
However, current works in this area are still developing and have yet to establish themselves.

Finally, the most direct method for improvement is to apply the system to other fabrics, such as other interlock types, injected samples, higher crimp fabrics or reinforcements with different yarn sizes.
This would add robustness to the networks and allow them to generalize to increasingly more complex and unseen cases such as full composite parts.
We are currently exploring this in order to better gauge the limits of this approach.

%% file: 8_appendix/main.tex
\section{Pseudo-CT generation from hand-drawn images}
\label{appendix:hand_drawn}
\input{1_hand_drawn}

\section{Visualization of learnt feature maps}
\label{appendix:features}
\input{2_features}

\section{Coherence through multiple slices}
\label{subsec:consistency}
\input{3_consistency}

%% file: 8_appendix/1_hand_drawn.tex
In order to further explore the generalization capabilities of the trained U-Net network (besides its use on the FE simulations), its responses for a collection of \enquote{hand-drawn} images is shown in~\cref{fig:unet_fake}.
All these figures were created using a vector graphics editor, however one could imagine using any textile modeling software (\eg WiseTex, TexGen) and a voxel conversion tool.
It is useful to highlight that these images are all varying in size and none of them have the same size as those used during training.
The convolution nature of the network allows for this flexibility.
However, it is important that the yarn in these images have a similar size to those used in the CT image (\ie \enquote{iso-resolution}).

\Cref{sfig:unet_fake_1,sfig:unet_fake_2} were created to inspect the isolated behavior of the network for both yarn types (\ie longitudinal or orthogonal).
\Cref{sfig:unet_fake_3,sfig:unet_fake_4} aim to test the generalization to other interlock types, orthogonal and through-the-thickness respectively, besides the layer-to-layer studied here.
\Cref{sfig:unet_fake_5} is a rough depiction of a \enquote{textile} that presents many undesirable issues.

\Cref{sfig:unet_fake_1_out,sfig:unet_fake_2_out} demonstrate that the network is not \enquote{used to} seeing the yarns isolated from the rest of the textile.
Hence, it fills the empty space with some traces of would-be yarns.
\Cref{sfig:unet_fake_3_out,sfig:unet_fake_4_out} present two interesting characteristics.
First, the \enquote{texture} for the longitudinal yarns presents clear discontinuities.
This is a sign that the network cannot handle these elevated curvatures (nonexistent in the ply-to-ply).
Second, the too-perfect positioning of the cross section yarns (\eg completely vertical columns) presents a curious challenge to the network.
It becomes clear that one of the contextual clues the network employs is the relative positioning of yarns.
Yet, when these remain constant, the network produces images for these yarns that are almost the same (see lower right of~\cref{sfig:unet_fake_4_out}).
\Cref{sfig:unet_fake_5_out} show that the network is capable of \enquote{accommodating} the longitudinal yarns so as to comply with the variable size of the cross section yarns.
Moreover, the two cases of contacting yarns present two different approaches: the first case (third layer) is to slightly separate the yarns, while the second one (fourth layer) calls for a more radical strategy, almost merging the yarns.
Clearly this is related to the cases of yarn inter-penetration that were present in the training phase.

Finally, a common quirk visible in all images is the addition of the film layer that was present on the sample during the acquisition (see bottom of~\cref{sfig:original_tomo}).
While this \enquote{feature} may seem innocuous at first, it does imply that some \emph{effort} during training was devoted to this banal aspect of the image.
This is a nice demonstration of one of the challenges related to deep learning, the at-times lack of control over what the network learns and what not.

\begin{figure}
    \centering
    \mysubfigure{0.06}{0.9}{figures/fake/1.png}{}{sfig:unet_fake_1}%
    \mysubfigure{0.13}{0.9}{figures/fake/2.png}{}{sfig:unet_fake_2}%
    \mysubfigure{0.12}{0.9}{figures/fake/3.png}{}{sfig:unet_fake_3}%
    \mysubfigure{0.19}{0.9}{figures/fake/4.png}{}{sfig:unet_fake_4}%
    \mysubfigure{0.30}{0.9}{figures/fake/5.png}{}{sfig:unet_fake_5}%
    \vspace*{\baselineskip}

    \mysubfigure{0.06}{0.9}{figures/fake/1_out.png}{}{sfig:unet_fake_1_out}%
    \mysubfigure{0.13}{0.9}{figures/fake/2_out.png}{}{sfig:unet_fake_2_out}%
    \mysubfigure{0.12}{0.9}{figures/fake/3_out.png}{}{sfig:unet_fake_3_out}%
    \mysubfigure{0.19}{0.9}{figures/fake/4_out.png}{}{sfig:unet_fake_4_out}%
    \mysubfigure{0.30}{0.9}{figures/fake/5_out.png}{}{sfig:unet_fake_5_out}%
    \caption{Qualitative evaluation of the U-Net generalization capabilities using
    \subref{sfig:unet_fake_1}-\subref{sfig:unet_fake_5} hand-drawn images, and
    \subref{sfig:unet_fake_1_out}-\subref{sfig:unet_fake_5_out} the corresponding pseudo-CT images}
    \label{fig:unet_fake}
\end{figure}

%% file: 8_appendix/2_features.tex
\begin{figure}
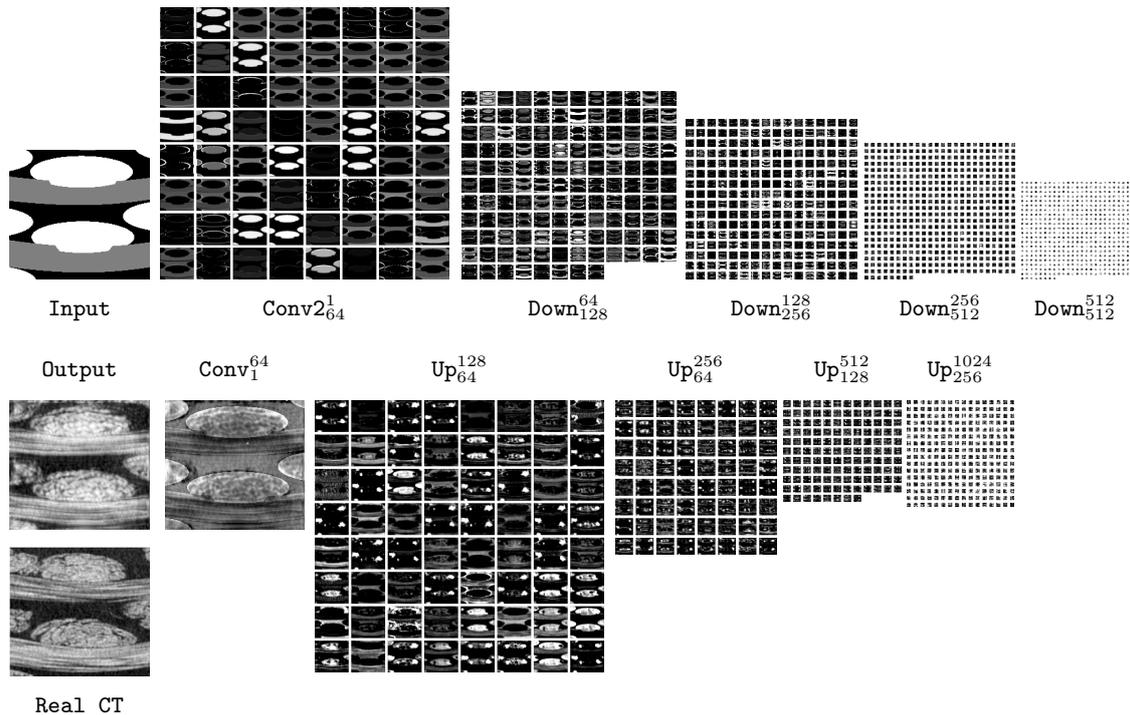

    \newcommand{\mapsb}[3]{%
        \begin{subfigure}[b]{#1\linewidth}%
        \centering\includegraphics[width=0.95\linewidth]{#2}\caption*{#3}%
        \end{subfigure}%
    }
    \newcommand{\mapst}[3]{%
        \begin{subfigure}[t]{#1\linewidth}%
        \centering\caption*{#3}\includegraphics[width=0.95\linewidth]{#2}%
        \end{subfigure}%
    }
    \centering
    \begin{subfigure}{0.9\linewidth}
        \mapsb{0.13}{figures/feature_maps/in.png}{$\texttt{Input}$}%
        \mapsb{0.27}{figures/feature_maps/inc.png}{$\texttt{Conv2}^{1}_{64}$}%
        \mapsb{0.20}{figures/feature_maps/down1.png}{$\texttt{Down}^{64}_{128}$}%
        \mapsb{0.16}{figures/feature_maps/down2.png}{$\texttt{Down}^{128}_{256}$}%
        \mapsb{0.14}{figures/feature_maps/down3.png}{$\texttt{Down}^{256}_{512}$}%
        \mapsb{0.10}{figures/feature_maps/down4.png}{$\texttt{Down}^{512}_{512}$}%

        \begin{subfigure}[t]{0.13\linewidth}
            \mapst{1.0}{figures/feature_maps/sigm.png}{$\texttt{Output}$}%
            \par\medskip
            \mapsb{1.0}{figures/feature_maps/out.png}{$\texttt{Real CT}$}%
        \end{subfigure}
        \mapst{0.13}{figures/feature_maps/outc.png}{$\texttt{Conv}^{64}_{1}$}%
        \mapst{0.27}{figures/feature_maps/up4.png}{$\texttt{Up}^{128}_{64}$}%
        \mapst{0.15}{figures/feature_maps/up3.png}{$\texttt{Up}^{256}_{64}$}%
        \mapst{0.11}{figures/feature_maps/up2.png}{$\texttt{Up}^{512}_{128}$}%
        \mapst{0.10}{figures/feature_maps/up1.png}{$\texttt{Up}^{1024}_{256}$}%
    \end{subfigure}
    \caption{\revision{Feature maps visualization of the U-Net architecture for a given input image,
    all inner layers are scaled by the same factor in order to highlight the change in image size}}%
    \label{fig:unet_activation_maps}
\end{figure}

In recent years, much effort has been dedicated to \enquote{explaining} the inner functioning of deep neural networks~\cite{Xie2020}.
However, the \emph{deep} nature of the model constitutes a challenge since the number of parameters in such a system are considerably many.
For example, the U-Net model contains over 9 million parameters that describe the convolutional filters that are applied to any input image.
While it is true that one could visualize these learnt filters as images, this would not be very useful in understanding what does each of them do.
However, it is possible to visualize the intermediate outputs that follow these filters, these are called \emph{feature maps}.
As such, the transformations that an input image undergoes (as it progresses through the network) will become apparent.
\revision{This strategy is chosen and the outputs of each block in the U-Net is shown in~\cref{fig:unet_activation_maps}, see~\cref{fig:unet_architecture} for reference on the convolutional blocks.}
The inverse relationship between size of the feature maps and their number becomes apparent.

%% file: 8_appendix/3_consistency.tex
One of the issues of the current 2.5D approach is that no consistency is imposed between consecutive slices.
As a consequence, when we seek to \enquote{gather} all results into a 3D space, some issues arise for the instance segmentation with Mask R-CNN (see~\cref{subsec:completion_missing_sections}). 
For completeness, we will explore the impact it has on the generation of pseudo-CT images (\ie the U-Net).

All YZ slices (orthogonal to warp direction) of the acquired CT volume $I_{CT}$ are fed to the U-Net and the outputs are stacked one after the other.
This effectively generates a new volume $I_{pCT}^\text{YZ}$ that can be observed in either the warp or the weft orientations (\ie YZ or XZ).
In the first case, we will observe the \emph{natural} response of the network.
While in the second one, the resulting slices show the consistency of the network through the volume.
The same can be done for the weft orientation so as to obtain $I_{pCT}^\text{XZ}$ with the equivalent orientations of interest XZ and YZ.
The central regions of the these volumes (\ie mid-slices) are shown in~\cref{fig:unet_consistency}.
A striking difference can be observed between the volumes seen along the generated direction (\ie~\cref{sfig:warp_unet_warp,sfig:weft_unet_weft}) versus those seen in the orthogonal one (\ie~\cref{sfig:warp_unet_weft,sfig:weft_unet_warp}).
However, the fact that the slices in~\cref{sfig:warp_unet_weft,sfig:weft_unet_warp} do not show many sharp gradients implies that, even if it is not enforced, the network does provide results that are  relatively consistent.
This highly desirable behavior implies that the network has learnt how to obtain contextual clues from the arrangement of yarns so that it knows how to \enquote{convert} the elliptical cross sections into realistic ones in a very consistent manner.
If this was not the case, this conversion would be almost random.

\begin{figure}
    \centering
    \mysubfigure{0.30}{0.85}{figures/consistency/warp_tomo.png}{Mid-YZ slice $x_{CT}$}{sfig:warp_tomo}%
    \mysubfigure{0.30}{0.85}{figures/consistency/warp_unet_warp.png}{Mid-YZ slice $x_{pCT}^\text{YZ}$}{sfig:warp_unet_warp}%
    \mysubfigure{0.30}{0.85}{figures/consistency/warp_unet_weft.png}{Mid-YZ slice $x_{pCT}^\text{XZ}$}{sfig:warp_unet_weft}%
    \vspace*{\baselineskip}

    \mysubfigure{0.30}{0.85}{figures/consistency/weft_tomo.png}{Mid-XZ slice $x_{CT}$}{sfig:weft_tomo}%
    \mysubfigure{0.30}{0.85}{figures/consistency/weft_unet_weft.png}{Mid-XZ slice $x_{pCT}^\text{XZ}$}{sfig:weft_unet_weft}%
    \mysubfigure{0.30}{0.85}{figures/consistency/weft_unet_warp.png}{Mid-XZ slice $x_{pCT}^\text{YZ}$}{sfig:weft_unet_warp}%
    \caption{Mid-slices for the three different volumes $I_{CT}$, $I_{pCT}^{YZ}$ and $I_{pCT}^{XZ}$, observed along
    \subref{sfig:warp_unet_warp}\subref{sfig:weft_unet_weft} the direction of inference, and
    \subref{sfig:warp_unet_weft}\subref{sfig:weft_unet_warp} along the orthogonal one 
    }
    \label{fig:unet_consistency}
\end{figure}